\documentclass[10pt,journal,compsoc]{IEEEtran}
\ifCLASSOPTIONcompsoc
  \usepackage[nocompress]{cite}
\else
  \usepackage{cite}
\fi
\ifCLASSINFOpdf
\else
\fi

\usepackage{microtype}
\usepackage{graphicx}
\usepackage{subfigure}
\usepackage{booktabs} 
\usepackage{hyperref}

\usepackage[utf8]{inputenc} 
\usepackage[T1]{fontenc}    
\usepackage{url}            
\usepackage{amsfonts}       
\usepackage{nicefrac}       
\usepackage{tablefootnote}
\usepackage{amssymb}
\usepackage{bm}
\usepackage{amsthm}
\usepackage{varwidth}
\usepackage{multirow}
\usepackage{makecell}
\usepackage{colortbl}
\usepackage{color}
\usepackage{algorithm}
\usepackage{algorithmic}
\usepackage[algo2e]{algorithm2e}

\usepackage{amsmath}
\usepackage{bbm}
\usepackage{extarrows}
\usepackage{enumitem} 
\usepackage{pdflscape}

\newtheorem{remark}{Remark}

\usepackage{xcolor}
\usepackage{xspace}
\usepackage{fixfoot}
\DeclareFixedFootnote*{\ftA}{When conditioning on the negative latent variable, we use the test samples with the ``smiling'' attribute. In terms of positive values, we use the test samples with the ``not smiling'' attribute.}
\DeclareFixedFootnote*{\ftRCGAN}{RCGAN fails to make a fine-grained translation w.r.t. the ``mouth'' attribute on CelebA-HQ dataset. So we did not collect its result.
}

\begin{document}

%
\title{TRIP: Refining Image-to-Image Translation via Rival Preferences}
%
%
%
%

\author{Yinghua Yao,
        Yuangang Pan,
        Ivor W. Tsang,~\IEEEmembership{Fellow,~IEEE,}
        and~Xin Yao,~\IEEEmembership{Fellow,~IEEE}
\thanks{
Email: yinghua.yao@student.uts.edu.au,
yuangang.pan@gmail.com,
ivor.tsang@uts.edu.au, 
xiny@sustech.edu.cn.}
\thanks{Main part of this work was done while the first author was at the University of Technology Sydney.}
}

\markboth{Journal of \LaTeX\ Class Files,~Vol.~14, No.~8, August~2015}%
{Shell \MakeLowercase{\textit{et al.}}: Bare Demo of IEEEtran.cls for Computer Society Journals}
%



\IEEEtitleabstractindextext{
\begin{abstract}
Relative attribute (RA), referring to the preference over two images on the strength of a specific attribute, can enable fine-grained image-to-image translation due to its rich semantic information. Existing work based on RAs however failed to reconcile the goal for fine-grained translation and the goal for high-quality generation. We propose a new model TRIP to coordinate these two goals for high-quality fine-grained translation. In particular, we simultaneously train two modules: a generator that translates an input image to the desired image with smooth subtle changes with respect to the interested attributes; and a ranker that ranks rival preferences consisting of the input image and the desired image. 
Rival preferences refer to the adversarial ranking process: (1) the ranker thinks no difference between the desired image and the input image in terms of the desired attributes; (2) the generator fools the ranker to believe that the desired image changes the attributes over the input image as desired. RAs over pairs of real images are introduced to guide the ranker to rank image pairs regarding the interested attributes only. With an effective ranker, the generator would ``win'' the adversarial game by producing high-quality images that present desired changes over the attributes compared to the input image. The experiments on two face image datasets and one shoe image dataset demonstrate that our TRIP achieves state-of-art results in generating high-fidelity images which exhibit smooth changes over the interested attributes.
\end{abstract}

}

\maketitle

\IEEEdisplaynontitleabstractindextext

%
\IEEEpeerreviewmaketitle

\begin{figure*}[!t]
    \centering
    \includegraphics[width=\linewidth]{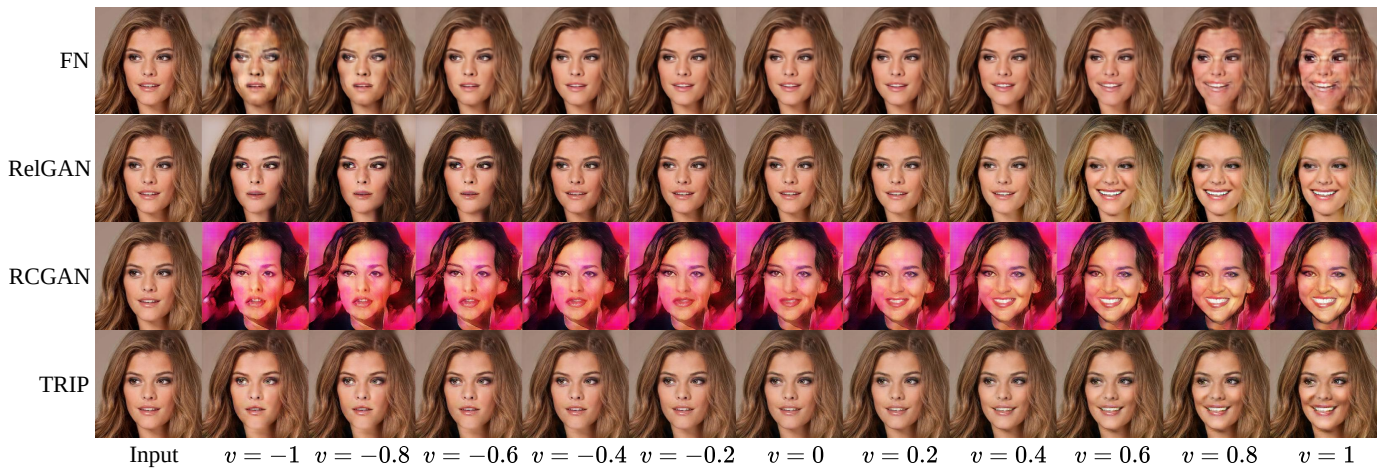}\vskip-0.1in
    \caption{\label{fig:celebahq_smile_inter}Comparison of fine-grained facial attribute (``smile'') translation on CelebA-HQ dataset. $v$ is a variable that controls the desired change of the ``smile'' attribute for the generated images.}
\end{figure*}

\IEEEraisesectionheading{\section{Introduction}\label{sec:introduction}}
\IEEEPARstart{I}{mage-to-image} (I2I) translation~\cite{isola2017image} aims to translate an input image into the desired ones with changes in some specific attributes.
Current literature can be classified into two categories: binary translation~\cite{zhu2017unpaired,kim2017learning,9367012}, e.g., translating an image from ``not smiling'' to ``smiling''; fine-grained translation~\cite{lample2017fader,he2019attgan,liu2018unified,DBLP:conf/bmvc/SaquilKH18}, e.g.,  generating a series of images  with smooth changes from ``not smiling'' to ``smiling''. 
In this work, we focus on the high-quality fine-grained I2I translation, namely, generating a series of realistic versions of the input image with smooth changes on the specific attributes (See Fig.~\ref{fig:celebahq_smile_inter}). Note that the desired high-quality images in our context are two folds: first, the generated images look as realistic as training images; second, the generated images \mbox{are only modified in terms of the specific attributes.}

Most existing methods~\cite{lample2017fader,liu2018unified,li2019latent,ding2020guided,liu2019conditional,he2019attgan,8886528} resort to binary attribute that indicates the presence (or absence) of a certain attribute in an image. However, their fine-grained I2I translation is unsatisfactory as the binary attributes have restrictive description capacity~\cite{wu2019relgan}. Instead, relative attribute (RA)~\cite{parikh2011relative}, referring to the preference of two images on the strength of the interested attribute, has more potentials in fine-grained translation because of its richer semantic information. For instance, for ``smiling'' attribute, it is coarse to simply classify whether an face image is either ``smiling'' or ``not smiling'' since there exist many images that are difficult to categorize. The difference in the ``smiling'' attribute between a pair of image can describe more subtle attribute changes, \mbox{such as the minor variation of the ``smiling'' angle.}

There are some works~\cite{DBLP:conf/bmvc/SaquilKH18,wu2019relgan} conducting fine-grained I2I translation based on continuous-valued RAs. All these methods proposed their model within the framework of generative adversarial networks (GANs)~\cite{goodfellow2014generative} that originally consists of a generator and a discriminator and basically designed two goals: the goal of interpolating RAs on generated images for fine-grained translation and the goal for good-quality generation.
However, all methods suffer from the conflict between two goals. Specifically, the methods model the RAs of real images by capturing the subtle difference over the attributes using an extra attribute model, e.g., a classification/ranking model, and count on the learned attribute model to generalize the learned attribute preferences to generated images with interpolated RAs. However, the learned attribute model never functions as expected since it generalizes poorly to the generated images with unseen RAs, which makes the generation deviate from the real data distribution. In a trade-off with the quality goal enforced by a discriminator, the generated images either fail to change smoothly over the interested attributes or suffer from low-quality issues.

In this paper, we propose a new adversarial ranking framework consisting of a ranker and a generator for high-quality fine-grained translation (TRIP). In particular, TRIP introduces the generated image into the training of the ranker that distills the discrepancy from the interested RAs. Since the supervision over the generated images is not accessible, we formulate the ranker into an adversarial ranking process using the rival preference, consisting of the generated image and the input image. Specifically, the ranker keeps ignoring the relative changes between the generated image and the input image regarding the interested attributes; while the generator aims to achieve the agreement from the ranker that the generated image holds the desired difference compared to the input. Competition between the ranker and the generator drives both two modules to improve themselves. Since our ranker can critic the generated image during its whole training process, and it no doubt can generalize well to generated images to ensure reliable fine-grained control over the target attributes.

We summarize our contributions as follows:
\begin{itemize}[leftmargin=0.2in]
    \item We propose Translation via RIval Preference (TRIP) consisting of a ranker and a generator for a high-quality fine-grained translation. The rival preference is constructed to evoke the adversarial training between the ranker and the generator, which enhances the ability of the ranker and encourages a better generator. 
    \item Our ranker enforces a continuous change between the generated image and the input image, which promotes a better fine-grained control over the interested attributes.
    \item Empirical results show that our TRIP achieves the state-of-art results on the fine-grained image-to-image translation task. Meanwhile, the input image can be manipulated linearly along the strength of the attributes.
    \item We further extend TRIP to the fine-grained I2I translation of multiple attributes. A case study demonstrates the efficacy of our TRIP in terms of disentangling multiple attributes and manipulating them simultaneously.
\end{itemize}

The rest of this paper is organized as follows. Section~\ref{sect:related works} reviews existing works related to fine-grained I2I translation and discusses their drawbacks. Section~\ref{sect:trip} presents the design principle and the whole framework of our proposed TRIP model in detail. Section~\ref{sect:experiments} empirically demonstrates the efficacy of TRIP on two face image datasets and one shoe image dataset. We conclude our paper in Section~\ref{sect:conclusion}.

\section{Related Works}
\label{sect:related works}
We review the literature related to fine-grained I2I translation.

\subsection{Based on Binary Attributes}
Many related works conduct fine-grained I2I translation based on coarse binary attributes. Though it is feasible to interpolate the attributes, the interpolation quality is unsatisfactory since the models are only trained on binary-valued attributes and thus the interpolation is ill-defined~\cite{berthelot2018understanding,wu2019relgan}. Apart from the defect of using binary attributes, the following works have other shortcomings inherited from the used generative models.

\textbf{AE-based} methods can provide a good latent representation of the input image. 
Some works~\cite{lample2017fader,liu2018unified,li2019latent,ding2020guided} proposed to disentangle the attribute-dependent latent variable from the image representation based on binary attributes but resorted to different disentanglement strategies.
Then the fine-grained translation can be derived by smoothly manipulating the attribute variable of the input image.
However, the reconstruction loss, which is used to ensure the image quality, cannot guarantee a high fidelity of the hallucinated images. 

\textbf{Flow-based} works proposed incorporating feature disentanglement mechanism into flow-based generative models~\cite{kondo2019flow}. However,  the designed multi-scale disentanglement requires large computation. \cite{liu2019conditional} applied an encoder to map the attribute space to the latent space of flow-based model by resorting to a binary attribute classifier, which fails to capture fine-grained attribution information.

\textbf{GAN-based} 
is a widely adopted framework for high-quality image generation.
\cite{he2019attgan,8886528} directly modeled the attributes with binary classification, which cannot capture detailed attribute information, and hence fail to make a smooth control over the attributes. 

\begin{figure*}[!t]
    \centering
    \includegraphics[width=\linewidth]{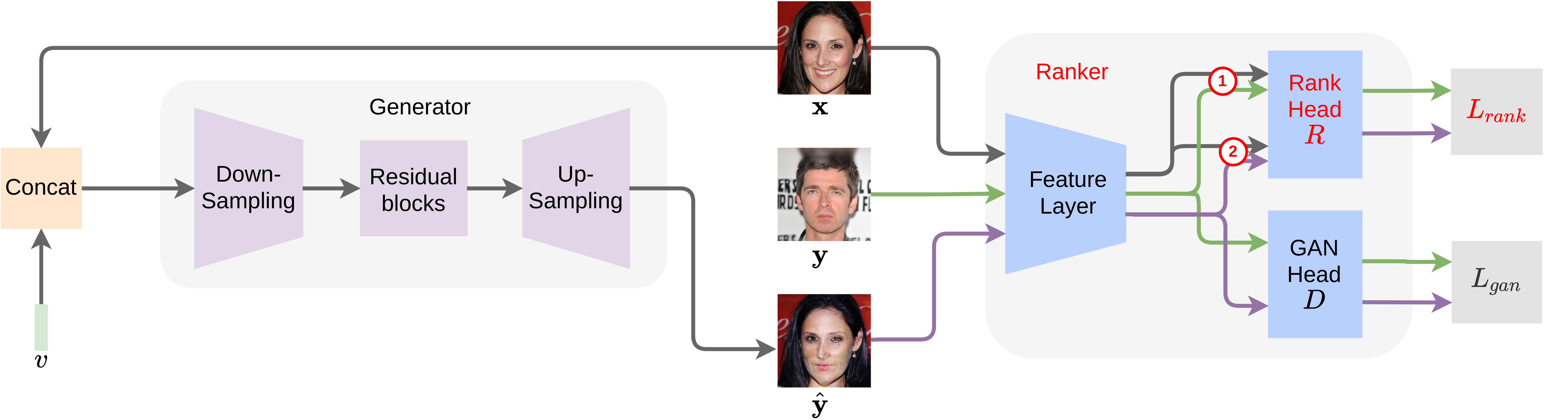}\vskip -0.1in
    \caption{\label{fig:whole_network}The network structure of TRIP. The main novelty is two folds: (1) the design of ranker and (2) the adversarial ranking process, which are denoted in red. \textcircled{1} and \textcircled{2} denote different image pairs, i.e., real image pairs and generated image pairs, corresponding to Fig.~\ref{fig:ranker} and Fig.~\ref{fig:trip}, respectively. $R$ and $D$ denotes the rank head and the GAN head, respectively.}
\end{figure*}

\subsection{Based on Relative Attributes}
Two works applied GAN as a base for fine-grained I2I translation through relative attributes, which can describe subtle changes on the interested attributes.
The main differences lie in the strategies of incorporating the preference over the attributes into the image generation process.
RCGAN~\cite{DBLP:conf/bmvc/SaquilKH18} adopts two critics consisting of a ranker, learning from the relative attributes of real images, and a discriminator, ensuring the image quality. Then the combination of two critics is aimed to guide the generator to produce high-quality fine-grained images while interpolating RAs on the generation.
However, since the ranker is only trained with real images and generalizes poorly to the unseen generated images, the ranker would induce the generator to generate out-of-data-distribution images which is opposite to the target of the discriminator, resulting in poor-quality images.
RelGAN~\cite{wu2019relgan} applies a matching-aware discriminator~\footnote{The matching-aware discriminator is modeled as a binary classifier as original GAN~\cite{goodfellow2014generative}. The difference is that the matching-aware discriminator distinguishes real triplets from fake triplets while the discriminator \mbox{in original GAN distinguishes real images from fake images.}}, which learns the joint data distribution of the triplet constructed with a pair of images and a discrete numerical label (i.e., relative attribute). In order to enable a smooth translation, RelGAN further interpolates the RAs on generated images while enforcing a constraint of interpolation quality. However, the constraint destroys smooth interpolation.

\subsection{Based on Other Priors}
\cite{deng2020disentangled} embedded 3D priors into adversarial learning. However, it relies on available priors for attributes, which limits the practicality. \cite{alharbi2020disentangled} proposed an unsupervised disentanglement method. It injects the structure noises to GAN for controlling specific parts of the generated images, which makes global or local features changed in a disentangled way. However, it is unclear how global or local features are related to facial attributes. Thus, it is difficult to change specific attributes.

\subsection{Our Model}
Our model is designed within the framework of GAN by utilizing relative attributes. Therefore, we regard the two most related work RCGAN~\cite{DBLP:conf/bmvc/SaquilKH18} and RelGAN~\cite{berthelot2018understanding} as our baselines. In addition, FN~\cite{lample2017fader}, a representative work of attribute disentanglement for fine-grained translation, is also regarded as our baselines. We  did  not  compare  with  AttGAN~\cite{he2019attgan} that enforces fine-grained translation via a binary classifier, since RelGAN outperforms AttGAN, which is shown in~\cite{wu2019relgan}.

\section{TRIP for Fine-grained I2I translation}
\label{sect:trip}
In this section, we propose a new model, named Translation via RIval Preferences (TRIP) for high-quality fine-grained image-to-image (I2I) translation by utilizing relative attributes (RAs), which learns a mapping that translates an input image to a set of realistic output images by smoothly controlling the specific attributes (Fig.~\ref{fig:celebahq_smile_inter}).
Previous works based on RAs, i.e., RCGAN and RelGAN, fail to reconcile the goal for fine-grained translation and the goal for high-quality generation, because their attribute model that is only trained with RAs on real images cannot generalize well to the generated images with unseen RAs. In contrast, our TRIP feeds RAs on real images as well as on generated images to model RAs and maintains rival preferences to coordinate the two goals.

We open our description with the one-attribute case. The whole structure of TRIP is shown in Fig.~\ref{fig:whole_network}, 
which consists of a generator and a ranker. The generator takes as input an image along with a continuous latent variable that controls the change of the attribute, and outputs the desired image; while the ranker delivers information in terms of image quality and the preference over the attribute to guide the learning of the generator. We implement the generator with a standard encoder-decoder architecture following~\cite{wu2019relgan}. In the following, we focus on describing the detailed design of the ranker and the principle behind it. Then we propose our TRIP model and extend it to the multiple-attribute case. Last, we compare our TRIP with RCGAN and RelGAN in technical details.

\subsection{Ranker for Relative Attributes}
\label{sec 3.1}
Relative attributes (RAs) are assumed to be most representative and most valid to describe the information related to the relative emphasis of the attribute,  owing to its simplicity and easy construction~\cite{parikh2011relative,DBLP:conf/bmvc/SaquilKH18}. For a pair of images $(\mathbf{x,y})$, RAs refer to their preference over the specific attribute: $\mathbf{y} \succ \mathbf{x}$ when $\mathbf{y}$ shows a greater strength than $\mathbf{x}$ on the target attribute and vice~versa. 

\begin{figure}[!htb]
    \centering
    \includegraphics[width=0.5\linewidth]{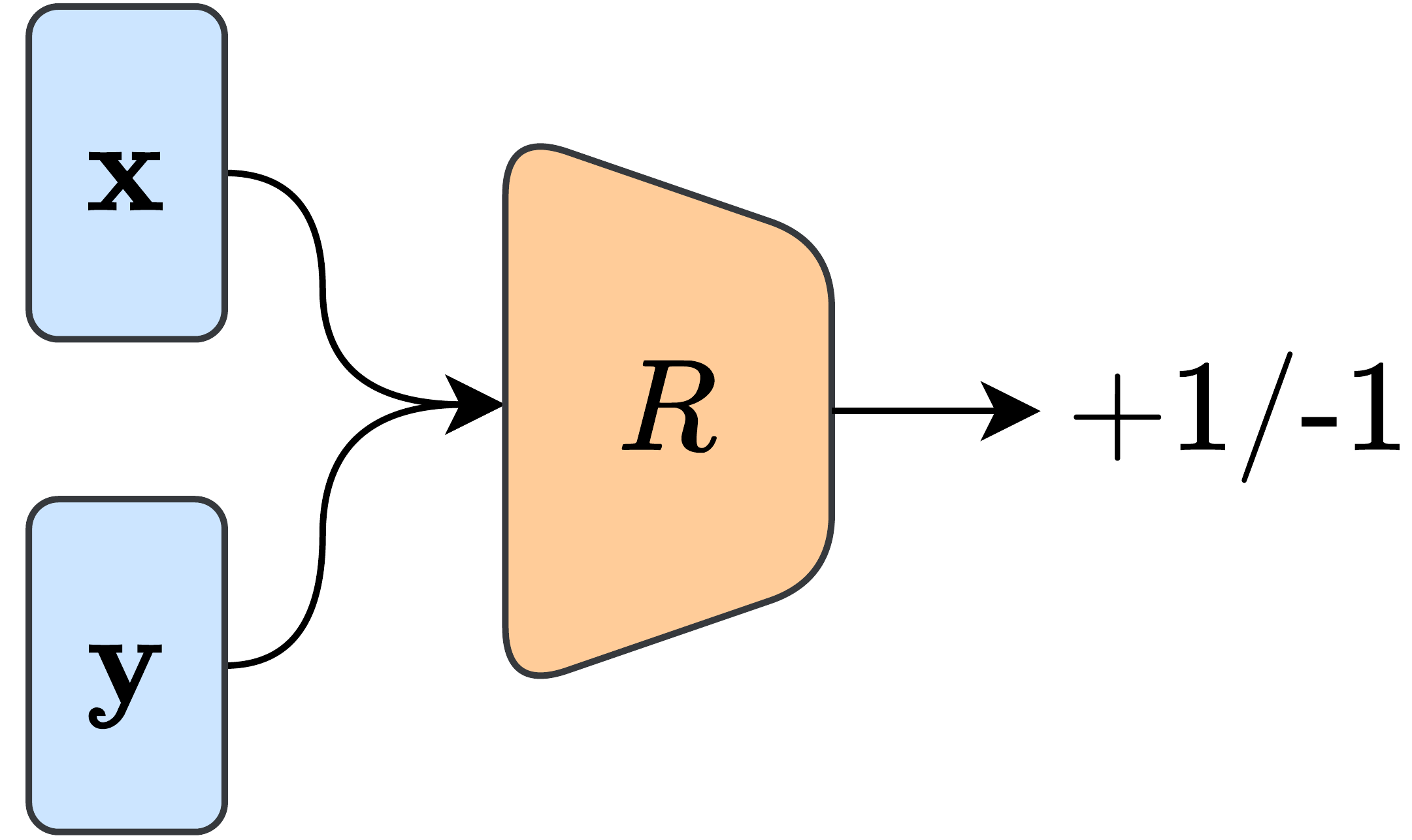}
    \caption{ \label{fig:ranker}The ranker model to learn relative attributes for real pairs}.
\end{figure}

Pairwise learning to rank is a widely-adopted technique to model the relative attributes~\cite{parikh2011relative}. Given a pair of images $(\mathbf{x,y})$ and its relative attribute, the pairwise learning to rank technique is formulated as a binary classification~\cite{cao2006adapting}, i.e.,
\begin{equation}
R(\mathbf{x,y})=\left\{\begin{array}{ll}
\ 1 & \mathbf{y} \succ \mathbf{x}; \\
-1 & \mathbf{y} \prec \mathbf{x},
\end{array}\right.
\label{eq:l2r}
\end{equation}
where $R(\mathbf{x,y})$ is the ranker's prediction for the pair of images $(\mathbf{x,y})$ (Fig.~\ref{fig:ranker}). 

Further, the attribute discrepancy between RAs, distilled by the ranker, can then be used to guide the generator to translate the input image into the desired one. 

However, the ranker is trained on the real image pairs, which only focuses on the modeling of preference over the attribute and ignores image quality. To achieve the agreement with the ranker, the generator possibly produces unrealistic images, which conflicts with the goal of the discriminator.

\subsection{Ranking Generalization by Rival Preferences}
\label{sec 3.2}

According to the above analysis, we consider incorporating the generated image pairs into the modeling of RAs, along with the real image pairs to reconcile the goal of the ranker and the discriminator. Consequently,  the resultant ranker will not only generalize well to the generated pairs but also avoid providing untrustworthy feedback by distinguishing the unrealistic pairs from the real pairs. 

Motivated by the adversarial training of GAN~\cite{goodfellow2014generative}, we introduce an adversarial ranking process between a ranker and a generator to incorporate the generated pairs into the training of ranker. To be specific,
\begin{itemize}[leftmargin=.2in]
\item \textbf{Ranker.} Inspired by semi-supervised GAN~\cite{odena2016semi}, we assign a pseudo label to the generated pairs. In order to avoid a biased influence on the ranking decision over real image pairs, i.e., positive (+1) or negative (-1), the pseudo label is designed to be zero. Note that a generated pair consists of a synthetic image and its input in order to connect the ranker \mbox{prediction to the latent variable.}
\begin{equation}
R(\mathbf{x},\Delta)=\left\{\begin{array}{ll}
+1 & (\Delta=\mathbf{y})  \wedge  (\mathbf{y} \succ \mathbf{x}); \\
-1 & (\Delta=\mathbf{y})  \wedge  (\mathbf{y} \prec \mathbf{x}); \\
\ 0 & \Delta=\hat{\mathbf{y}}.
\end{array}\right.
\label{eq:d}
\end{equation}
where $\hat{\mathbf{y}}$ denotes the output of the generator given the input $\mathbf{x}$ and an attribute variable $v$, i.e., $\hat{\mathbf{y}}=G(\mathbf{x}, v)$. In particular, $v$ controls the desired change of attributes for the generated images. $\Delta$ is a placeholder that can be either a real image $\mathbf{y}$ or a generated image~$\hat{\mathbf{y}}$.
\item \textbf{Generator.} The goal of the generator is to achieve the consistency between the ranking prediction $R(\mathbf{x}, \hat{\mathbf{y}})$ and the corresponding latent variable $v$. 
When $v>0$, the ranker is supposed to believe that the generated image $\hat{\mathbf{y}}$ has a larger strength of the specific attribute than the input $\mathbf{x}$, i.e., $R(\mathbf{x}, \hat{\mathbf{y}})=+1$; and vice versa.
\begin{equation}
R(\mathbf{x},\hat{\mathbf{y}})=\left\{\begin{array}{ll}
+1 &  v>0; \\
-1 &  v<0.
\end{array}\right.
\label{eq:g_bin}
\end{equation}
\end{itemize}

\begin{figure}[!ht]
    \centering
    \includegraphics[width=0.9\linewidth]{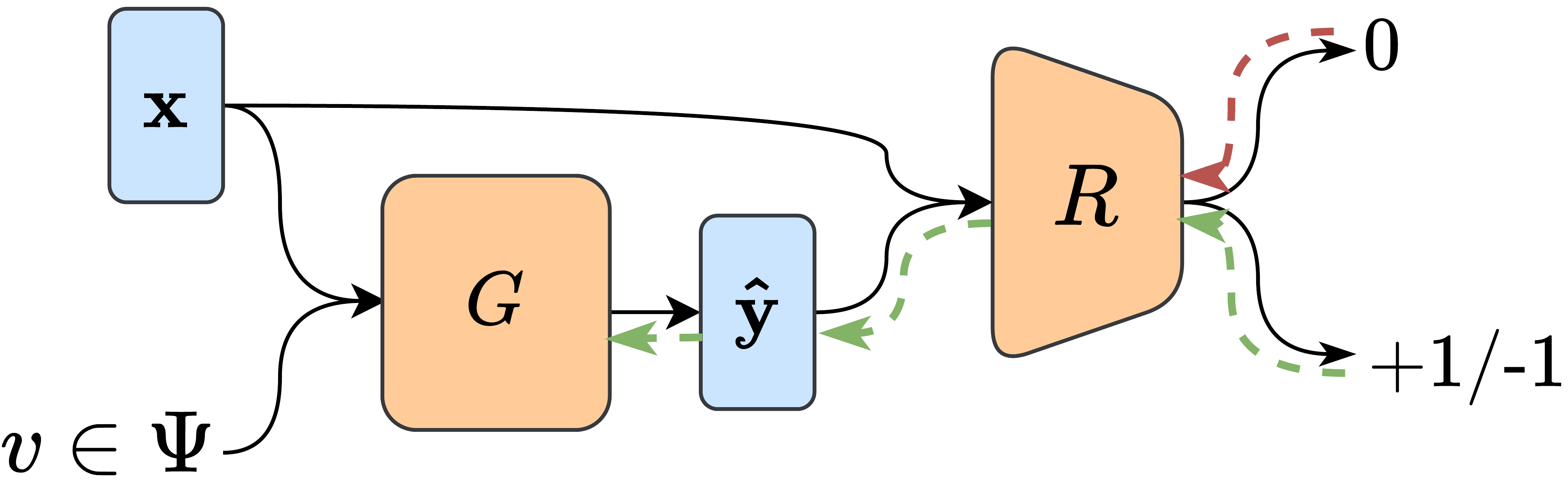}
    \caption{\label{fig:trip}Rival Preferences for the generated image pairs between the ranker and the generator, which ensure the ranker generalizes well to the generated pairs. $\Psi$ denotes $[\text{-}1,0)\cup(0,1]$.}
\end{figure}

We denominate the opposite goals between the ranker and the generator w.r.t. the generated pairs as \textbf{rival preferences}\footnote{``rival'' means adversarial. We use it to distinguish it from adversarial training in the community.}. Such preferences extend the adversarial game defined on binary classification~\cite{goodfellow2014generative} to ranking. An intuitive example of the rival preference is given in Fig.~\ref{fig:trip} for better understanding. 

The ranker is promoted in terms of the following aspects: (1) The function of the ranker on the real image pairs is not changed. The generated pairs are uniformly sampled regarding their latent variables. By assigning label zero, the ranking information implied within the pairs is neutralized to maintain the ranking performance on the real image pairs. 
(2) The ranker avoids providing biased ranking prediction for unrealistic image pairs. As we constrain the generated pairs at the decision boundary, i.e, $R(\mathbf{x}, \hat{\mathbf{y}})=0$, the ranker is invariant to the features specified by the generated pairs~\cite{chapelle2008analysis}, suppressing the influence of the unrealistic features on the ranking decision. 
(3) The ranker can capture the exclusive difference over the specific attribute through the adversarial process. Since the ranker rejects to give effective feedback for unrealistic image pairs, only the realistic image pairs can attract the attention of the ranker.  Therefore, the ranker only passes the effective information related to the target attribute to the generator.

Then, we introduce a parallel head following the feature layer to ensure the image quality together with a rank head, shown in Fig.~\ref{fig:whole_network}. According to the above analysis, the ranker would not evoke conflicts with the goal of the image quality. Therefore, we successfully reconcile the two goals of image quality and the extraction of the attribute difference.
With a powerful ranker, the generator would ``win'' the adversarial game by producing the realistic pairs consistent with the latent variable.

\begin{remark}[Assigning zero to similar real image pairs]
It is natural to assign zero to pairs $\{(\mathbf{x,y})|\mathbf{y} = \mathbf{x}\}$, where $=$ denotes that $\mathbf{x}$ and $\mathbf{y}$ have the same strength in the interested attribute.
They can improve the ranking prediction~\cite{10.1145/1390334.1390382}. 
\end{remark}

\subsection{Linearizing Ranking Output for Smooth Translation}\label{sec 3.3}
Eq.~\eqref{eq:g_bin} models the relative attributes of the generated pairs as a binary classification,  which would fail to enable a fine-grained translation since the subtle changes implied by the latent variable are not distinguished by the ranker. For example, given $v_1>v_2>0$, the ranker gives same feedbacks ($\text{+}1$) for $(\mathbf{x}, \hat{\mathbf{y}}_1)$ and  $(\mathbf{x},\hat{\mathbf{y}}_2)$, which loses the discrimination between the two pairs.
To achieve the fine-grained translation, we linearize the ranker's output for the generated pairs so as to align the ranker prediction with the latent variable. We thus reformulate the binary classification as the regression:
\begin{equation}
    R(\mathbf{x},\hat{\mathbf{y}}) = v.  
    \label{eq:g_continuous}
\end{equation}
Given two latent variables $1>v_2>v_1>0$, the ranking prediction for the pair generated from $v_2$ should be larger than that from $v_1$, i.e., $1>R(\mathbf{x}, \hat{\mathbf{y}}_2)>R(\mathbf{x},\hat{\mathbf{y}}_1)>0$.
The ranker’s outputs for the generated pairs would be linearly correlated to the corresponding latent variable. 
Since the ranker's output for a generated pair reflects the difference between the generated image and the input image, the generated image can change smoothly over the input image according to the latent variable.

\begin{figure*}[!htb]
	\begin{minipage}{0.325\linewidth}
	\centerline{\includegraphics[width=\textwidth]{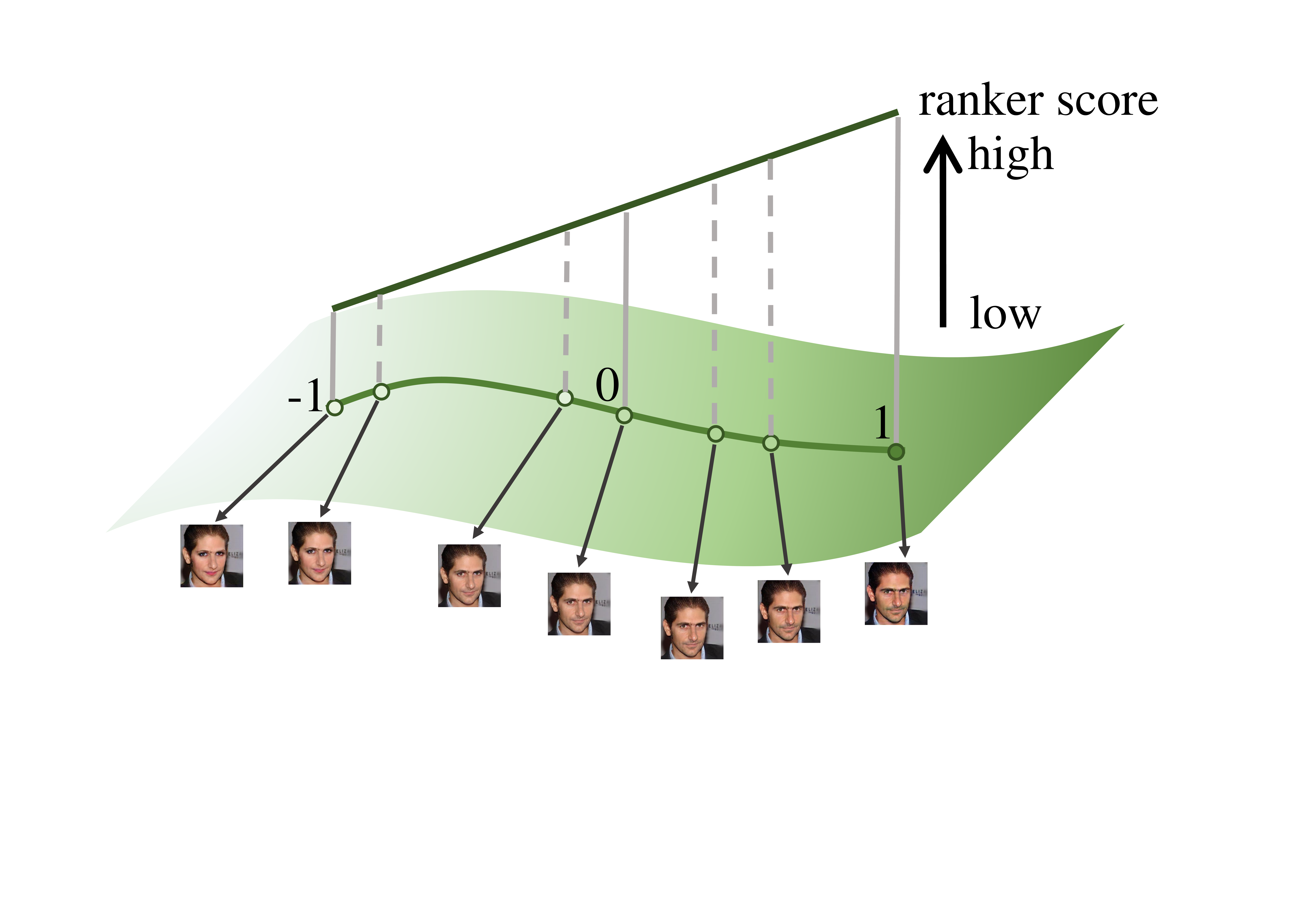}}
	\centerline{(a) TRIP}
	\end{minipage}
	\begin{minipage}{0.325\linewidth}
	\centerline{\includegraphics[width=\textwidth]{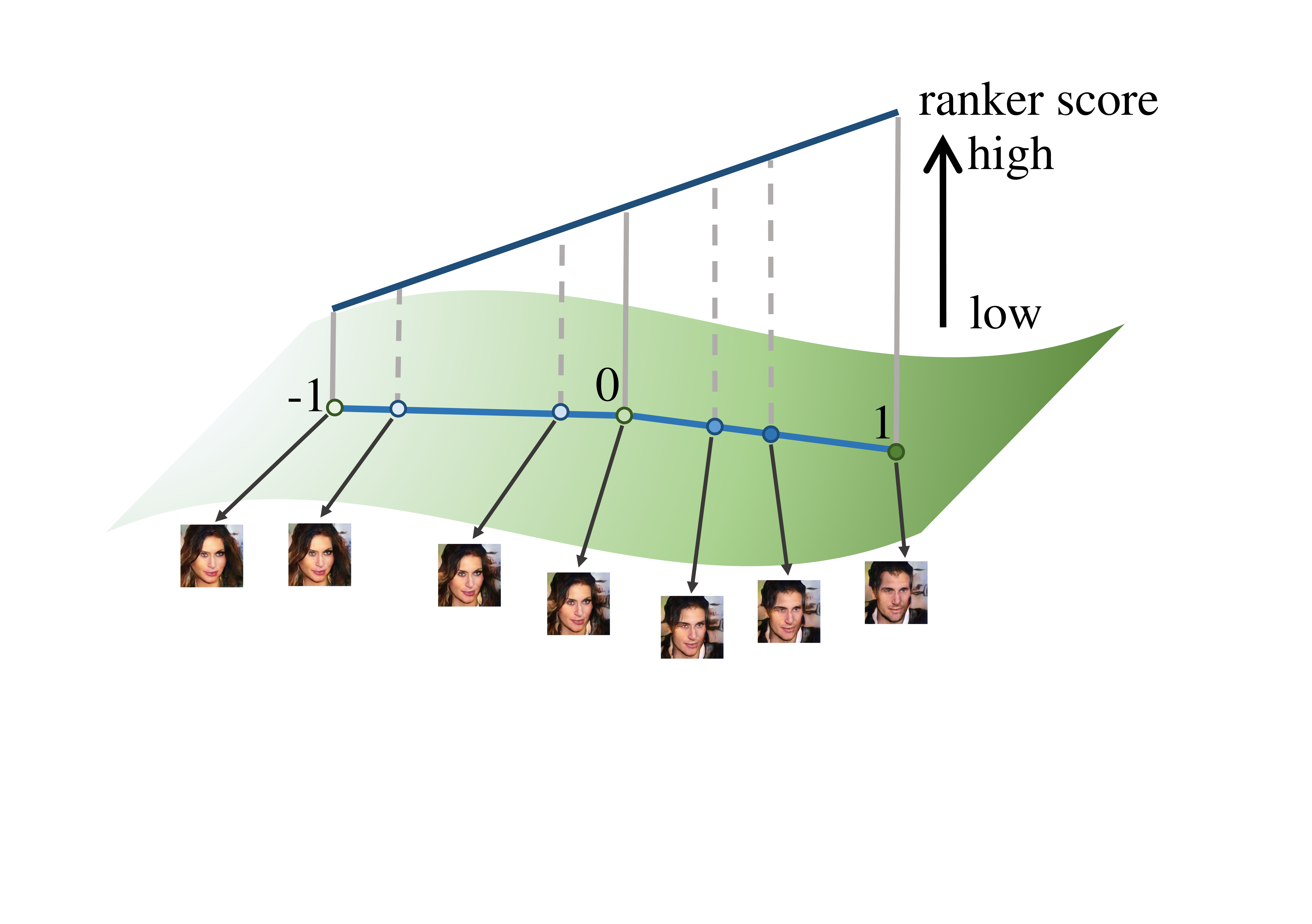}}
	\centerline{(b) RCGAN}
	\end{minipage}
	\begin{minipage}{0.35\linewidth}
	\centering
	\centerline{\includegraphics[width=\textwidth]{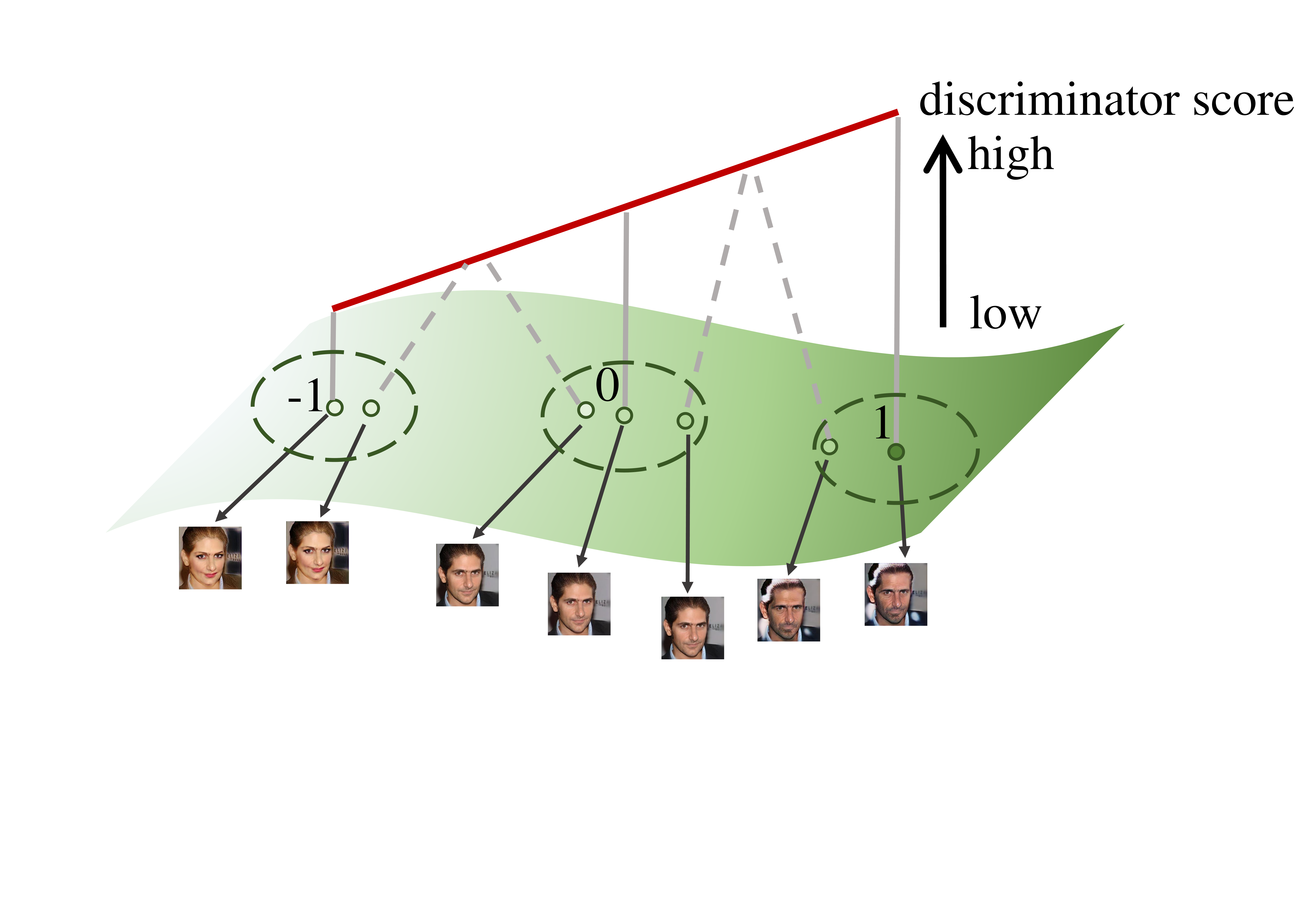}}
	\centerline{(c) RelGAN}
	\end{minipage}		
  \\  
  \caption{\label{fig:trip_rcgan_relgan} Correspondence between the ranker and the generated images by TRIP, RCGAN, and RelGAN, respectively. \textbf{TRIP:} the generated images (deep green line) lie on the data manifold (green curved surface), which have high quality. Meanwhile, they are linearly correlated to the ranker scores, which delivers smooth translation. \textbf{RCGAN:} the generated images (blue line, as distinct from the color of data manifold) are out of the data manifold, although exhibiting linear correlation with the ranker output. \textbf{RelGAN:} the generated images gather on the data main manifold within three green circles, and fail to spread out linearly with its discriminator score.}
\end{figure*}
\subsection{Translation via RIval Preferences (TRIP)}
\label{sec 3.4}
In the following, we introduce the loss functions for the two parallel heads in the ranker. The overall network structure can be seen in Fig.~\ref{fig:whole_network}.

\textbf{Loss of rank head $R$:} we adopt the least square loss for the ranking predictions. The loss function for the rank head and the generator is defined as:
\begin{subequations}
\begin{align}
    L^{R}_{rank} &= \mathbb{E}_{p(\mathbf{x,y},r)} \left[\left(R\left(\mathbf{x,y}\right)-r\right)^2\right] \label{eq:trip_ls_d}\\
    &\qquad\qquad +\lambda \mathbb{E}_{p(\mathbf{x})p(v)}  \left[\left(R\left(\mathbf{x},G(\mathbf{x}, v)\right)-0\right)^2\right];\nonumber \\
    L^{G}_{rank} &= \mathbb{E}_{p(\mathbf{x})p(v)}  \left[\left(R\left(\mathbf{x},G(\mathbf{x}, v)\right)-v\right)^2\right]\label{eq:trip_ls_g},
\end{align}
\label{eq:trip_rank}
\end{subequations}
where \begin{small}$r =\begin{cases} \ 1 & \mathbf{y} \succ \mathbf{x} \\ \ 0 & \mathbf{y} = \mathbf{x}\\-1 & \mathbf{y} \prec \mathbf{x} \end{cases}$\end{small} denotes the relative attribute. $p(\mathbf{x,y},r)$ are the joint distribution of real image preferences. $\hat{\mathbf{y}}=G(\mathbf{x}, v)$. $p(\mathbf{x})$ is the distribution of the training images. $p(v)$ is a uniform distribution $[-1,1]$. $\lambda$ is the weight factor that determines the strength of adversarial training between the ranker and the generator.

By optimizing $L_{rank}^R$ (Eq.~\eqref{eq:trip_ls_d}), the ranker is trained to predict correct labels for real image pairs and assign zero for generated pairs, i.e., Eq.~\eqref{eq:d}. By optimizing $L_{rank}^G$ (Eq.~\eqref{eq:trip_ls_g}), the generator is trained to output the desire image $\hat{\mathbf{y}}$, 
where the difference between $\hat{\mathbf{y}}$ and $\mathbf{x}$ is consistent with the latent variable $v$, i.e., Eq.~\eqref{eq:g_continuous}. Therefore, the ranker learns to distill the discrepancy from the interested RAs. Meanwhile, the two rival goals on the generated pairs raise an adversarial training between the ranker and the generator. That is, the generator urges the ranker to predict that there are desired changes in the generated images while the ranker predicts that there are no changes. The competitive game promotes the improvement of two modules simultaneously.

\textbf{Loss of GAN head $D$:} to be consistent with the above rank head and also ensure a stable training, a regular least square GAN's loss is adopted:
\begin{subequations}
\begin{align}
    L^{D}_{gan} &= \mathbb{E}_{p(\mathbf{x})} \left[\left(D\left(\mathbf{x}\right)-1\right)^2\right]\label{eq:gan_d} \\
    & \qquad\qquad\qquad + \mathbb{E}_{p(\mathbf{x})p(v)} \left[\left(D\left(G(\mathbf{x}, v)\right)-0\right)^2\right];\nonumber \\
    L^{G}_{gan} &= \mathbb{E}_{p(\mathbf{x})p(v)} \left[\left(D\left(G(\mathbf{x}, v)\right)-1\right)^2\right]\label{eq:gan_g},
\end{align}
\label{eq:trip_gan}
\end{subequations}
where $1$ denotes the real image label while $0$ denotes the fake image label. 

Jointly training the rank head and the GAN head, the gradients backpropagate through the shared feature layer to the generator. Then our TRIP can conduct the high-quality fine-grained I2I translation.

\subsection{Extended to the Multiple Attributes}
\label{sec 3.5}
To generalize TRIP to multiple ($K$) attributes, we use vectors $\mathbf{v}$ and $\mathbf{r}$ with $K$ dimension to denote the latent variable and the preference label, respectively. Each dimension controls the change of one interested attribute. In particular, the ranker consists of one GAN head and $K$ parallel rank head. The overall loss function is summarized as follows: 
\begin{subequations}
\begin{align}
    L^{R}_{rank} &= \mathbb{E}_{p(\mathbf{x,y},\mathbf{r})} \sum_k \left[\left(R_k\left(\mathbf{x,y}\right)-
    \mathbf{r}_k\right)^2\right]] \label{eq:trip_ls_d_k}\\
    & \quad\quad +\lambda \mathbb{E}_{p(\mathbf{x})p(\mathbf{v})} \sum_k \left[\left(R_k\left(\mathbf{x},G(\mathbf{x}, \mathbf{v})\right)-0\right)^2\right]; \nonumber\\
    L^{G}_{rank} &= \mathbb{E}_{p(\mathbf{x})p(\mathbf{v})} \sum_k \left[\left(R_k\left(\mathbf{x},G(\mathbf{x}, \mathbf{v})\right)-\mathbf{v}_k\right)^2\right]\label{eq:trip_ls_g_k},
\end{align}
\end{subequations}
where $R_k$ is the output of the $k$-th rank head. $\mathbf{v}_k$ and  $\mathbf{r}_k$ are the $k$-th dimension of $\mathbf{v}$ and $\mathbf{r}$,  respectively.

\subsection{Discussion on TRIP from a Technical Perspective}
In this section, we analyze the advantages of TRIP in technical details. For better understanding, we compare TRIP with the two most related state-of-art works, i.e., RCGAN and RelGAN In particular, we plot the correspondence between the generated pairs (consisting of the generated images and the coupled input images) and the ranker score in Fig.~\ref{fig:trip_rcgan_relgan}. For simplicity of presentation, we omit the input images. 
\label{sect:comparison}

\subsubsection{TRIP Compared with RCGAN} 
\label{sect:tripVSrcgan}
\textbf{Ranker for RA.} TRIP and RCGAN both apply a ranker to explicitly model relative attributes, namely, mapping the difference between a pair of images to its corresponding RA labels. The ranker can guide the translation with a specific change in terms of the interested attributes. For example, when targeting the ``smiling'' attribute, the generated images are only modified in the related facial region, like the ``smiling'' angle. In addition, as seen in Fig.~\ref{fig:trip_rcgan_relgan}a and Fig.~\ref{fig:trip_rcgan_relgan}b, the rankers of TRIP and RCGAN both output scores linearly with the desired change of generated images over the input images.

\textbf{Adversarial ranking VS. Ranking.} Our ranker reconciles the goal for fine-grained translation and the goal for high-quality generation via adversarial ranking (See Fig.~\ref{fig:trip_rcgan_relgan}a). The ranker in RCGAN only models RAs of real images for guiding a fine-grained translation but conflicts with the discriminator in RCGAN for high-quality generation (See Fig.~\ref{fig:trip_rcgan_relgan}b). Thus RCGAN has extremely poor generation, which will be verified in the experiment section.

\subsubsection{TRIP Compared with RelGAN} 
\label{sect:tripVSrelgan}
\textbf{Ranker VS. Matching-aware discriminator.} TRIP distills the attribute difference by modeling RAs with a ranking model. In contrast, RelGAN resorts to a matching-aware discriminator to learn attribute information. Specifically, the discriminator judges whether an input-output pair matches the relative attributes or not with binary classification. Namely, the pair $(\mathbf{x, y})$ along with its ground-truth RA label $r$ is classified as $1$ while the pair $(\mathbf{x, y})$ along with the wrong RA label $r$ is classified as $0$. A ranker is more effective to capture subtle differences exclusively regarding the target attributes compared to a binary classifier. 

\textbf{Adversarial ranking VS. Interpolation.} TRIP introduces an adversarial ranking process, ensuring that the ranker can criticize the generated images with unseen RAs during its training process. Thus, TRIP can perform fine-grained control over the (ranker-guided) generator regarding the target attributes. Instead, RelGAN resorts to interpolation over the discrete RAs to generalize the attribute difference to unseen generated images, which is ill-defined [1] and needs extra constraints (interpolation loss, which will be discussed in the following). 

\textbf{Maintaining the linear tendency or not.}
TRIP: recalling Eq.~\eqref{eq:g_continuous}, the ranker’s outputs for the generated pairs
is linearly correlated to the corresponding latent variable (See Fig.~\ref{fig:trip_rcgan_relgan}a).  So the generated image $G(\mathbf{x}, v)$ can change smoothly over the input image $\mathbf{x}$ according to the RA. 
\mbox{RelGAN: recalling its interpolation loss in Sect 3.5 of ~\cite{wu2019relgan},}
\begin{equation*}
    \begin{aligned}
        \min_{G}\ \mathbb{E}_{p(\mathbf{x})p(r)p(\alpha)} [\|D_{\textrm{Interp}}(G(\mathbf{x}, \alpha r))\|^{2}], 
    \end{aligned}
\end{equation*}
where $r \in \{-1,0,1\}$ and $\alpha \in [0,1]$. $D_{\textrm{Interp}}$ recovers the interpolation ratio and assigns zero to non-interpolated images. Such interpolation loss aims at forcing interpolated images to be indistinguishable from non-interpolated images, which is designed for high-quality interpolation but destroys smooth translation (See Fig.~\ref{fig:trip_rcgan_relgan}c). Experiment results verify our claim.

\begin{figure*}[!t]
    \centering
    \includegraphics[width=\linewidth]{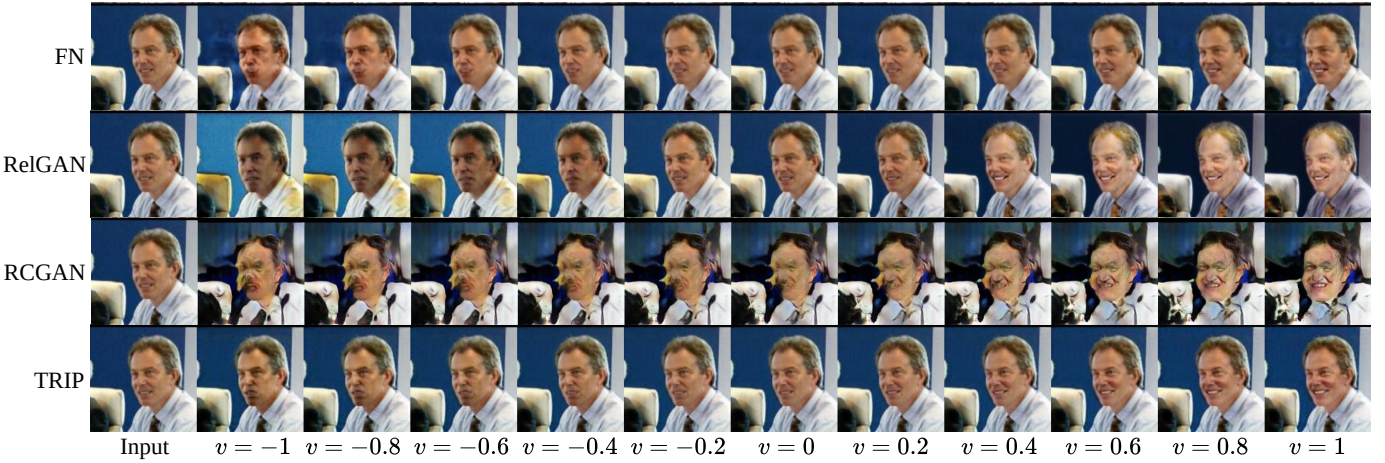}\vskip-0.15in
    \caption{\label{fig:lfw_smile_inter} Comparison of fine-grained facial attribute (``smile'') translation on LFWA dataset.}
\end{figure*}

\begin{figure*}[!htb]
    \centering
    \includegraphics[width=\linewidth]{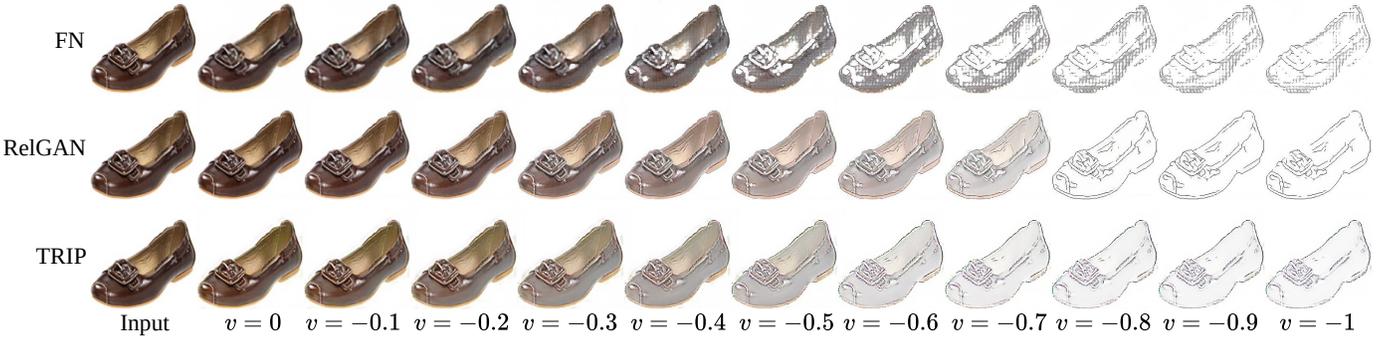}\vskip-0.15in
    \caption{\label{fig:utzap_shoe2edge_inter} Comparison of fine-grained translation (``shoe$\rightarrow$edge'')  on UT-Zap50K dataset.}
\end{figure*}
\section{Experiments}
\label{sect:experiments}
In this section, we compare our TRIP with various baselines on the task of fine-grained I2I translation. We first verify that our ranker can distinguish the subtle difference in a pair of images. Thus we propose to apply our ranker for evaluating the fine-graininess of image pairs generated by various methods. We finally extend TRIP to the translation of multiple attributes.

\textbf{Datasets.} We conduct experiments on two face image datasets, i.e., the high quality subset of Celeb Faces Attributes Dataset (CelebA-HQ)~\cite{DBLP:conf/iclr/KarrasALL18} and Labeled Faces in the Wild with attributes (LFWA)~\cite{liu2015faceattributes}, and one shoe image dataset, i.e., UT-Zappos50K dataset (UT-Zap50K)~\cite{finegrained}. CelebA-HQ consists of $30$K face images of celebrities with $40$ binary attributes. LFWA has $13,143$ images with $73$ binary attributes. UT-Zap50K contains $50,025$ catalog shoe images together with corresponding edge images~\cite{isola2017image}, where the binary label is 1 when an image is a shoe image and is 0 when an image is an edge image. 
We resize the images to $256 \times 256$ for three datasets.
The relative attributes are obtained for any two images $\mathbf{x}$ and $\mathbf{y}$ based on the binary labels. Thus we can make a fair comparison with other baselines in terms of same supervision information. For instance, for ``smile’’ attribute, we construct the comparison $\mathbf{x}>\mathbf{y}$ when the ``smile'' label of $\mathbf{x}$ is 1 while the ``smile'' label of $\mathbf{y}$ is 0, and vice versa. Note that to enhance the fine-grained translation on UT-Zap50K, we augment the training images by mixing up shoe images with the corresponding edge images~\cite{DBLP:conf/iclr/ZhangCDL18}.  Specifically, a mix-up image is obtained by $(1-\gamma)\mathbf{x}+\gamma\mathbf{y}$, where $\mathbf{y}$ is a shoe image and $\mathbf{x}$ is $\mathbf{y}$'s corresponding edge image. The corresponding mix-up label is $\gamma$, where $\gamma \in (0, 1)$. 

\textbf{Implementation Details.} 
As the translation is conducted on the unpaired setting, the cycle consistency loss $L_{cycle}$ are usually introduced to keep the identity of faces when translation~\cite{zhu2017unpaired,wu2019relgan,he2019attgan}. An orthogonal loss $L_o$ and the gradient penalty loss $L_{gp}$ are added to stabilize the training following~\cite{wu2019relgan}. The weighting factor for $L_{gan}$, $L_{cycle}$, $L_{o}$ and $L_{gp}$ are $\lambda_g$, $\lambda_{c}$, $\lambda_o$ and $\lambda_{gp}$, respectively. Except $\lambda_g=0.5$ for CelebA-HQ, $\lambda_g=5$ for LFWA and  $\lambda_g=2.5$ for LFWA, we set the same parameter for all datasets. Specifically, we set $\lambda=0.5$, $\lambda_c=2.5, \lambda_{gp}=150, \lambda_o=10^{-6}$. We use the Adam optimizer [23] with $\beta_1 = 0.5$ and $\beta_2 = 0.999$.
The learning rate is set to $1e\text{-}5$ for the ranker and $5e\text{-}5$ for the generator. The batch size is set to $4$. 
We split the dataset into training/test with a ratio $90/10$ for face datasets. We use 40K shoe-edge image pairs (80K images in total) and the corresponding mix-up images (40K images in total) as training data and $10,025$ shoe-edge image pairs ($20,050$ images in total) as test data for UT-ZAP50K dataset.
We pretrain our TRIP only with $L_{gan}$ to enable a good reconstruction for the generator. By doing so, we ease the training by sequencing the learning of our TRIP. That is, we first make a generation with good quality. Then when our GAN begins to train, the ranker can mainly focus on the relationship between the generated pairs and its corresponding conditional $v$, rather than handling the translation quality and the generation quality together. All the experiment results are obtained by a single run. Each run contains 100K iterations.

\textbf{Baselines.} We compare TRIP with FN~\cite{lample2017fader}, RelGAN~\cite{wu2019relgan} and RCGAN~\cite{DBLP:conf/bmvc/SaquilKH18}. We use the released codes of FN, RelGAN and RCGAN.

\textbf{Evaluation Metrics.}
Following~\cite{wu2019relgan}, we use three metrics to quantitatively evaluate the performance of fine-grained translation. 
Standard deviation of structural similarity (SSIM) measures the performance of fine-grained translation. Mean square error (MSE) and  Frechet Inception Distance (FID) measure the visual quality for shoe dataset and face datasets, respectively. 
Accuracy of Attribute Swapping (AAS) evaluates the accuracy of the binary image translation. The swapping for the attribute is to translate an image, e.g., from ``smiling'' into ``not smiling''.
\begin{itemize}[leftmargin=.15in]
    \item \textbf{SSIM.} We first apply the generator to produce a set of fine-grained output images $\{\mathbf{x}_1, \ldots, \mathbf{x}_{l}\}$ by conditioning an input image and a set of latent variable values from $-1$ to $1$ with a step $0.2$. Then $l=\frac{1-(-1)+0.2}{0.2}=11$. We compute \textbf{the standard deviation of the structural similarity (SSIM)}~\cite{wang2004image} between $x_{i-1}$ and $x_i$ as follows:
    \begin{equation}
    \sigma\left(\left\{\operatorname{SSIM}\left(\mathbf{x}_{i-1}, \mathbf{x}_{i}\right) \mid i=1, \cdots, 11\right\}\right).
    \end{equation}
    We calculate SSIM for each image from the test dataset and average them to get the final score. Smaller SSIM denotes better smooth translation.
    \item \textbf{MSE.} For shoe$\rightarrow$edge in UT-ZAP50K, the input-output examples are paired. Namely, the input image (shoe) and the output image (edge) only differ in their style, i.e., photo style and edge style, respectively. Then the mean square error between the translated images from generative models and the ground-truth output images can be calculated to measure the visual quality of translation. We conduct translation for all images in UT-ZAP50K dataset, obtaining $50,025$, and calculate the MSE. Smaller MSE denotes generation with better quality.
    \item \textbf{FID.} As the input-ouptut examples in face datasets are unpaired, we evaluate the statistical difference between the translated images and the training dataset to measure the visual quality. It is calculated with $30K$ translated images on CelebA-HQ dataset and $13,143$ translated images on LFWA dataset. Smaller FID denotes generation with better quality.
    \item \textbf{AAS.} The accuracy is evaluated by a facial attribute classifier that uses the Resnet-18 architecture~\cite{he2016deep}. To obtain AAS, we first translate the test images with the trained GANs and then apply the classifier to evaluate the classification accuracy of the translated images coupled with its swapping attribute. Higher accuracy means that more images are translated as desired.
\end{itemize}

\begin{figure*}[!tb]
    \centering
    \includegraphics[width=\linewidth]{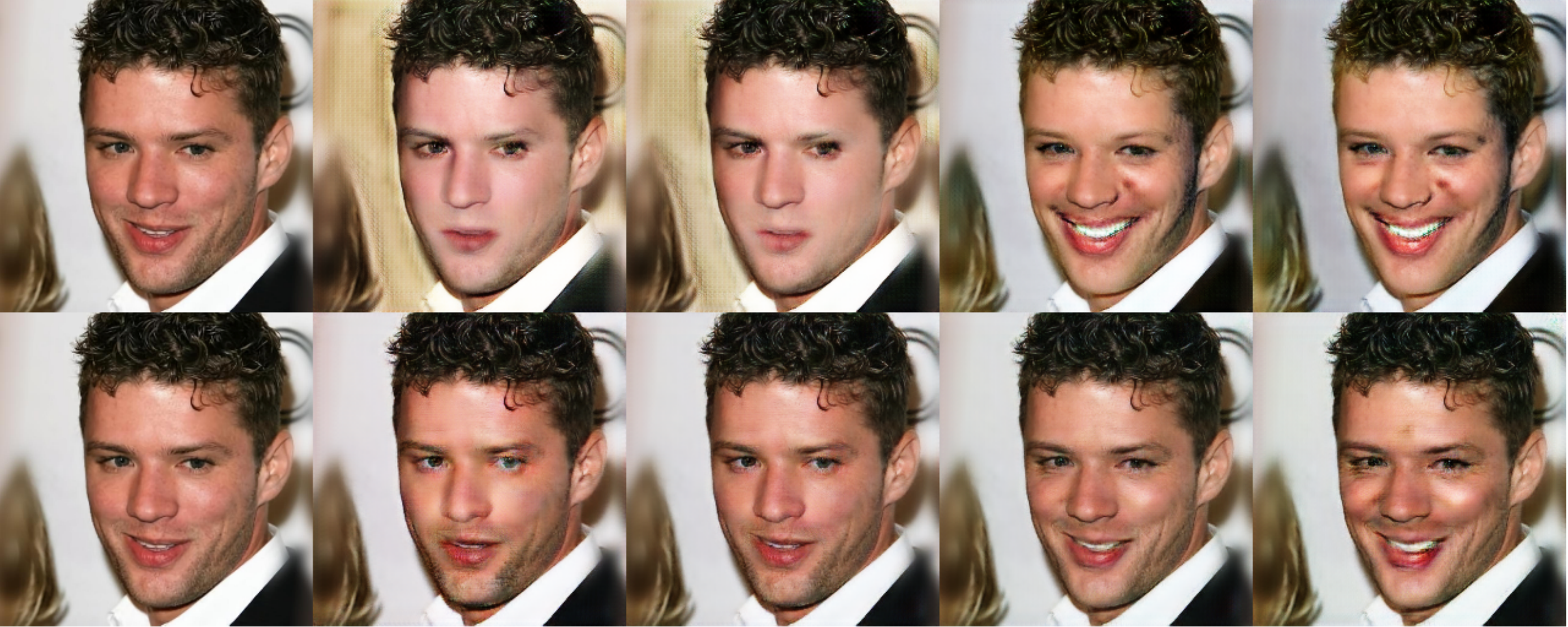}\vskip-0.1in
    \caption{\label{fig:celebhq_big_image}Image comparison of TRIP (second row) with RelGAN (first row) w.r.t. the ``smile'' attribute. The 1st column is input image. Other columns are generated images conditioning on $v=-1,-0.5,0.5,1$ from left to right.}
\end{figure*}

\begin{figure*}[!t]
    \centering
    \includegraphics[width=\linewidth]{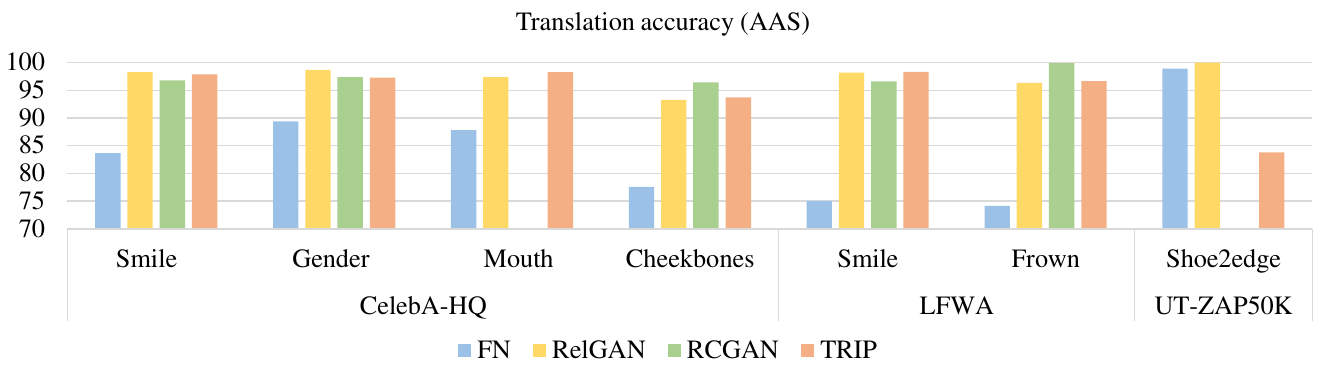}\vskip-0.1in
    \caption{\label{fig:aas} Translation accuracy (AAS, higher is better) of FN, RCGAN, RelGAN and TRIP on CelebA-HQ, LFWA and UT-ZAP50K. RCGAN fails to make fine-grained translations w.r.t. the ``mouth'' attribute on CelebA-HQ  and ``shoe$\rightarrow$edge'' on UT-ZAP50K. So we do not collect their results.}
\end{figure*}

\begin{table*}[!htb]
\Large
\centering
\caption{\label{tb:iq} Fine-grained performance (SSIM) and image quality (FID/MSE) of FN, RCGAN, RelGAN and TRIP on CelebA-HQ, LFWA and UT-ZAP50K. The best results are highlighted in bold. Considering the value range, we round four decimal places for SSIM and round two decimal places for FID and MSE. RCGAN fails to make fine-grained translations w.r.t. the ``mouth'' attribute on CelebA-HQ  and ``shoe$\rightarrow$edge'' on UT-ZAP50K. So we do not collect their results.}
\renewcommand{\arraystretch}{1.2}
\setlength{\tabcolsep}{0.6mm}{	
\scalebox{0.59}{
\begin{tabular}{c|cccc|cc|c|cccc|cc|c}
\toprule[1.3pt]
\multirow{3}{*}{Model} 
& \multicolumn{7}{c|}{Fine-grained (SSIM)}                  & \multicolumn{7}{c}{Image quality (FID/MSE)}               \\ \cline{2-15}
& \multicolumn{4}{c|}{CelebA-HQ} & \multicolumn{2}{c|}{LFWA}  &UT-Zap50K  
& \multicolumn{4}{c|}{CelebA-HQ (FID)}                       &\multicolumn{2}{c|}{LFWA (FID)}  &  UT-Zap50K (MSE)  \\ 
& Smile & Gender & Mouth & Cheekbones  & Smiling & Frown    & shoe$\rightarrow$edge 
& Smile & Gender & Mouth & Cheekbones & Smiling & Frown     & shoe$\rightarrow$edge \\ \midrule[1pt]
FN  & 0.0122 & \textbf{0.0036} & 0.0075 & 0.0039 & 0.0066   & 0.0049 & 0.0395  & 41.48 & 48.66 & 42.79  & 43.15         & \textbf{12.59} & \textbf{11.37} & 6987.76 \\
RCGAN  & 0.0084 & 0.0138  & -  & 0.0099 &  0.0079 & 0.0106  & - & 398.05 & 418.06  & -  & 385.27  & 437.93 & 425.59 &     -  \\
RelGAN  & 0.0512 & 0.0924  & 0.0261 & 0.0510 & 0.0137          & 0.0159 & 0.0023  & 10.92  & 31.29 & 9.55  & 10.28         & 16.76  & 16.21 & 6973.37  \\
TRIP & \textbf{0.0030} & 0.0077 & \textbf{0.0017} & \textbf{0.0028} &  \textbf{0.0008} & \textbf{0.0005} & \textbf{0.0010} &  \textbf{10.19} & \textbf{26.47} & \textbf{7.18} & \textbf{9.50} &  22.25  & 23.65  & \textbf{6187.88} \\
\toprule[1.3pt]
\end{tabular}}}
\end{table*}

\subsection{Fine-grained Image-to-Image Translation}
We conduct fine-grained I2I translation on a single attribute. On CelebA-HQ dataset, we translate images in terms of ``smile'', ``gender'', ``mouth open'' and ``high cheekbones'' attributes, respectively.  On LFWA dataset, we translate images in terms of ``smile'' and ``Frown'' attributes, respectively. On UT-ZAP50K dataset, we translate the shoe images to edge images, i.e., ``shoe$\rightarrow$edge''.
We show that our TRIP achieves the best performance on the fine-grained I2I translation task comparing with various strong baselines in the following three metrics.

\textbf{Visual results.} Fig.~\ref{fig:celebahq_smile_inter}, Fig.~\ref{fig:celebhq_big_image} and Fig.~\ref{fig:lfw_smile_inter} show that: (1) all methods can translate the input image into ``more smiling'' when $v>0$ or ``less smiling'' when $v<0$ on CelebA-HQ and LFWA, respectively. The degree of changes is consistent with the numerical value of $v$. (2) Our TRIP's generation achieves the best visual quality, generating realistic output images that are different from the input images only in the specific attribute. In contrast, FN suffers from image distortion issues. RelGAN's generation not only changes the specific attribute ``smile'', but is also influenced by other irrelevant attributes, e.g., ``hair color''. RCGAN exhibits extremely poor generation results. Fig.~\ref{fig:utzap_shoe2edge_inter} shows that: FN and TRIP can translate shoe images into edge images smoothly but RelGAN fails. 

\textbf{Fine-grained score.} We present the quantitative evaluation of the fine-grained translation in Table~\ref{tb:iq}. Our TRIP achieves the lowest SSIM scores for three datasets, consistent with the visual results. Note that a trivial case to obtain a low SSIM is when the translation is failed. Namely, the generator would output the same value no matter what the latent variable is. Therefore, we further apply AAS to evaluate the I2I translation in a binary manner. RelGAN, RCGAN and TRIP mostly achieve over $95\%$ accuracy except for FN (seen in Fig.~\ref{fig:aas}. Under this condition, it guarantees that a low SSIM indeed indicates the output images change smoothly with the latent variable. 

\textbf{Image quality score.} Table~\ref{tb:iq} presents the quantitative evaluation of the image quality. (1) Our TRIP achieves the best image quality with the lowest FID scores. (2) FN achieves the best FID on LFWA dataset. Because the FN achieves a relatively low accuracy of the translation, $<75\%$ in Fig.~\ref{fig:aas}, many generated images would be the same as the input image. It means that the statistics of the translated images are similar to that of the input images, leading to a low FID. (3) RCGAN has the worst FID scores, consistent with the visual results in Fig.~\ref{fig:celebahq_smile_inter} and Fig.~\ref{fig:lfw_smile_inter}.

\subsection{Physical Meaning of Ranker Output}
\label{sect:ranker_meaning}
\begin{figure}[!b]
    \centering
    \includegraphics[width=.72\linewidth]{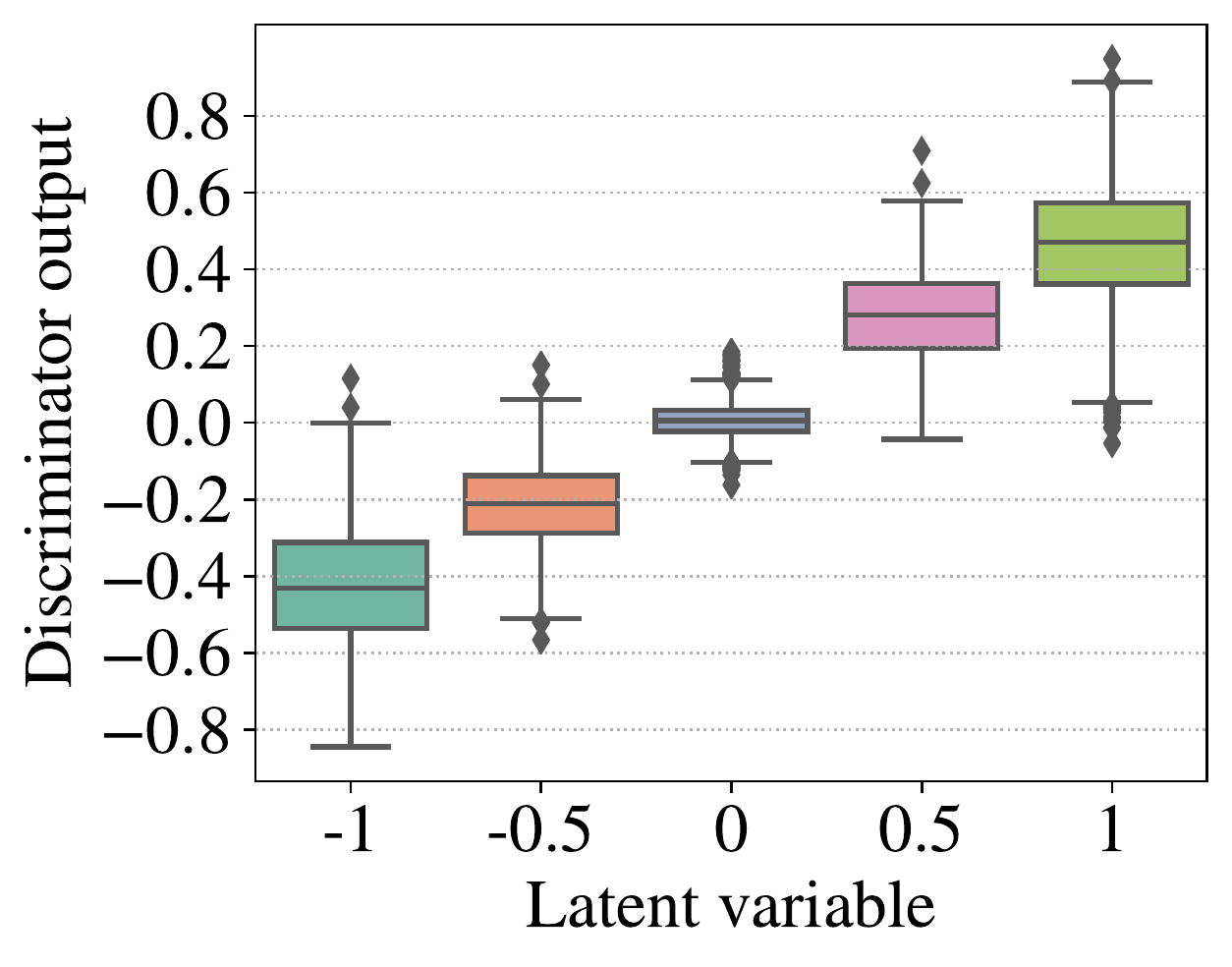}\vskip -0.1in
    \caption{\label{fig:d_inter} The box plot of the ranker's output for generated pairs with different values of the latent variable.}
\end{figure}

\begin{figure}[!htb]
    \centering
 \begin{minipage}{0.235\textwidth}
  \centering
    \includegraphics[width=\linewidth]{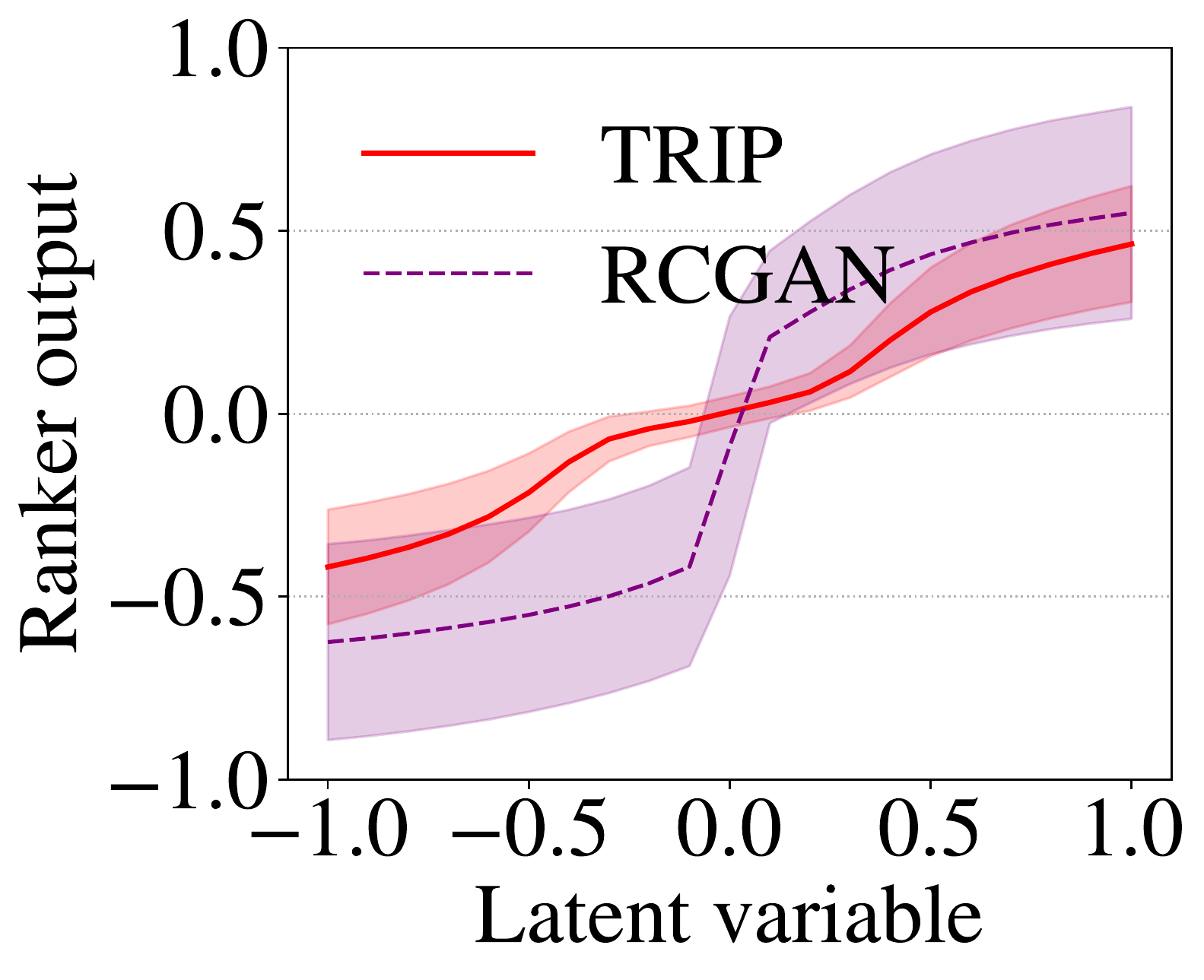}
  \end{minipage}
  \begin{minipage}{0.235\textwidth}
  \centering
    \includegraphics[width=\linewidth]{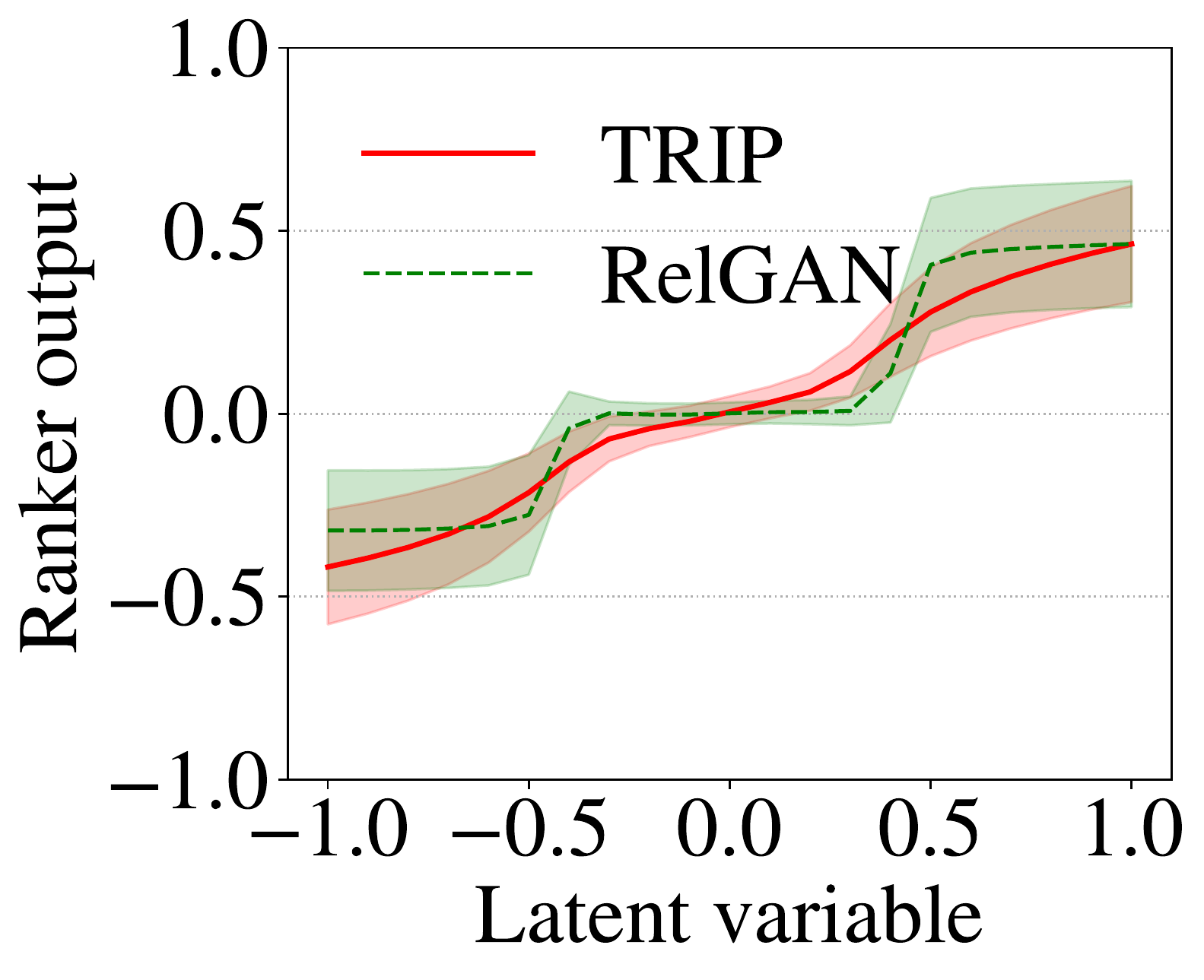}
  \end{minipage}
  \begin{minipage}{0.24\textwidth}
  \centering
    \includegraphics[width=\linewidth]{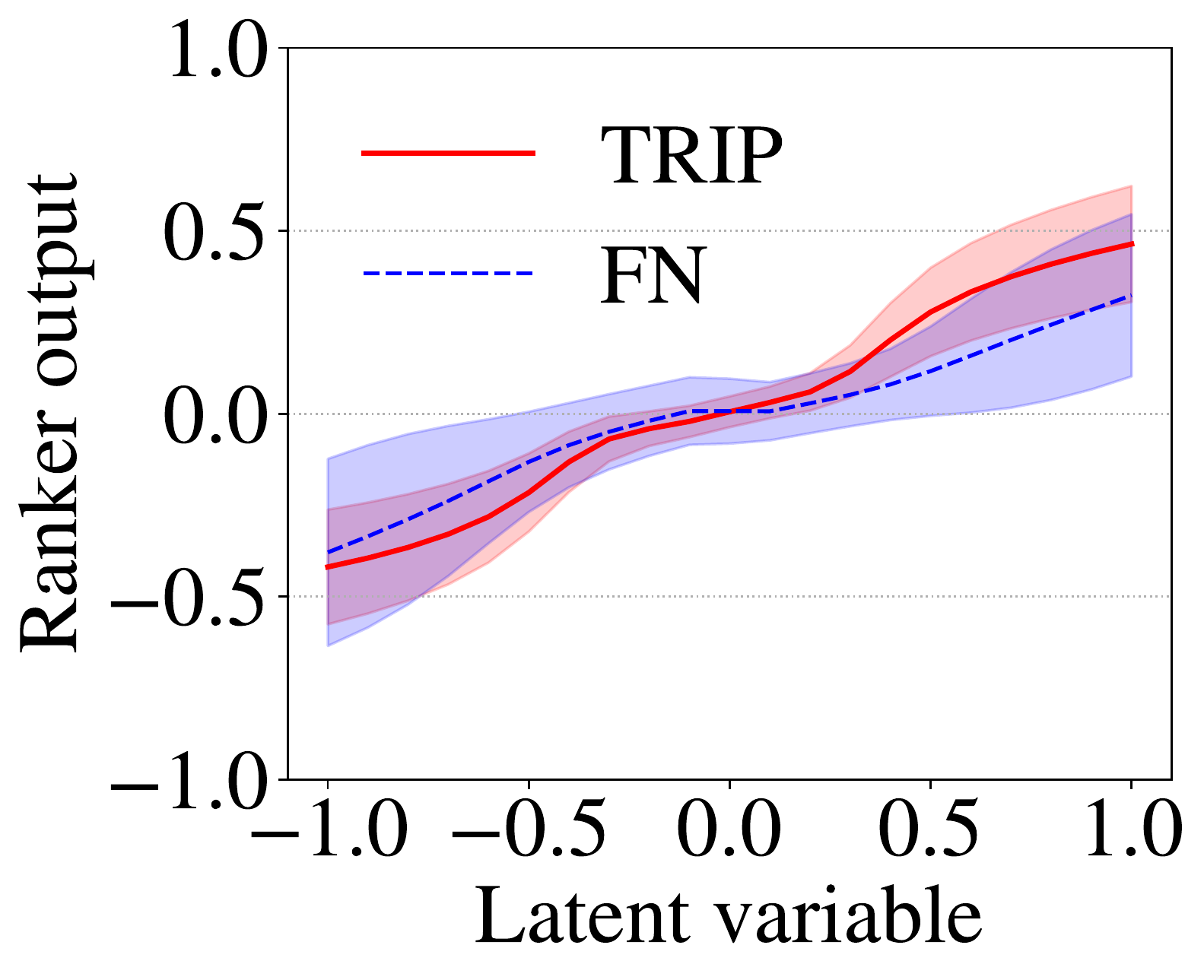}
  \end{minipage}
  \begin{minipage}{0.24\textwidth}
  \centering
  \renewcommand{\arraystretch}{1.4}
	\setlength{\tabcolsep}{.8mm}{	
		\scalebox{0.65}{
\begin{tabular}{ccccc}
\toprule[1.3pt]
\multicolumn{1}{c}{$v$} & RelGAN           & RCGAN            & FN      & TRIP             \\ \hline
$-1.$                   & $0.165$          & $0.268$          & $0.256$ & $\mathbf{0.157}$          \\
$-0.5$                  & $0.164$          & $0.265$          & $0.137$ & $\mathbf{0.107}$          \\
$0.$                    & $\mathbf{0.030}$          & $0.354$          & $0.089$ & ${0.042}$ \\
$0.5$                   & $0.183$          & $0.273$          & $0.122$ & $\mathbf{0.121}$          \\
$1.$                    & ${0.172}$ & ${0.290}$ & $0.222$ & $\mathbf{0.159}$ \\ 
\toprule[1.3pt]
\end{tabular}}}
  \end{minipage}
  \vskip-0.1in \caption{\label{fig:d_line} The first three subfigures plot the ranker's output for generated pairs in terms of different latent variables. The curve shows the mean of the output, while the shaded region depicts the standard deviation of the output. We summarize the standard deviation in the table for better understanding.} 
\end{figure}

\begin{figure*}[!t]
    \centering
    \includegraphics[width=0.9\linewidth]{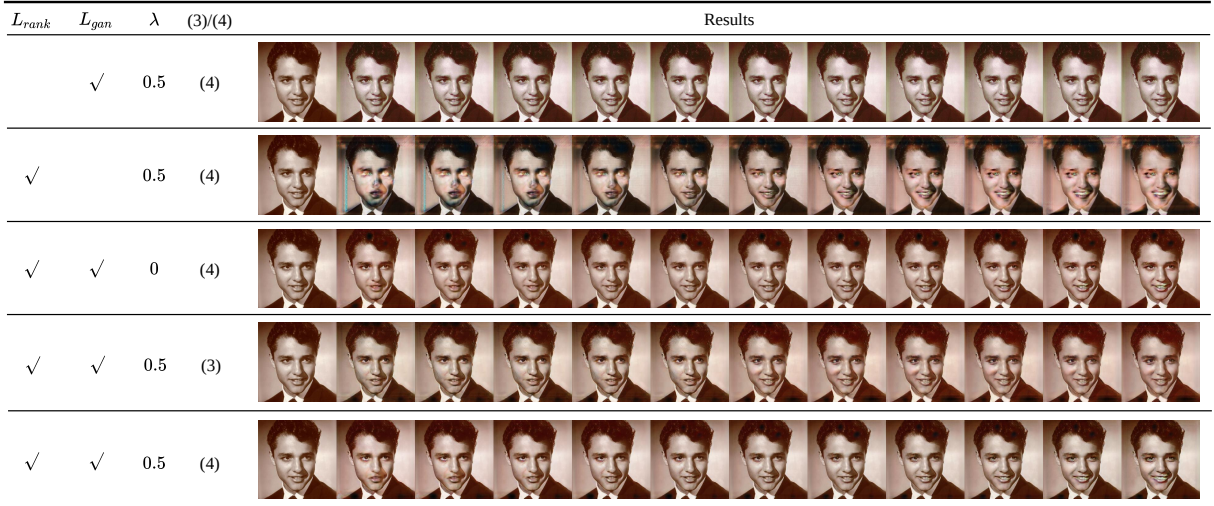}
    \caption{ \label{fig:ablation} Fine-grained I2I translation on CelebA-HQ (``smile'' attribute). The first column is the input while others are generated conditioned on $v$ from $-1$ to $1$. $L_{rank}$, $L_{gan}$, (3) and (4) refer to Eq.~\eqref{eq:trip_rank}, Eq.~\eqref{eq:trip_gan}, Eq.~\eqref{eq:g_bin}, and Eq.~\eqref{eq:g_continuous}, respectively.} 
\end{figure*}

As shown in Fig.~\ref{fig:d_inter}, (1) for a large $v$, the ranker would output a large prediction. It demonstrates that the ranker indeed generalizes to synthetic imaged pairs and can discriminate the subtle change among each image pair. 
(2) The ranker can capture the whole ordering instead of the exact value w.r.t. the latent variable. Because the ranker that assigns $0$ to the generated pairs inhibits the generator's loss optimizing to zero, although our generator's objective is to ensure the ranker output values are consistent with the latent variable. However, the adversarial training would help the ranker to achieve an equilibrium with the generator when convergence, so that the ranker can maintain the whole ordering regarding the latent variable.

From Fig.~\ref{fig:celebahq_smile_inter} and Table~\ref{tb:iq}, it shows that when conditioning on different latent variables, our TRIP can translate an input image into a series of output images that exhibit the corresponding changes over the attribute. We then evaluate the function of our ranker using these fine-grained generated pairs. It verifies that our ranker's output well-aligns to the relative change in the pair of images.

We further evaluate fine-grained I2I translations w.r.t. the ``smile'' attribute on the test dataset of CelebA-HQ (Fig.~\ref{fig:d_inter}).
The trained generator is applied to generate a set of $G(\mathbf{x},v)$ by taking as inputs an image $x$ and $v=-1.0,-0.5,0.0,0.5,1.0$, respectively. Note we use the test samples with the ``smiling'' attribute for a negative $v$, or the test samples with the ``not smiling'' attribute otherwise. Then we collect the output of the ranker for each generated pair and plot the density in terms of different~$v$.

\subsection{Linear Tendency on the Latent Variable}
\begin{figure}[!tb]
	\begin{minipage}{0.32\linewidth}
    \centerline{\includegraphics[width=\textwidth]{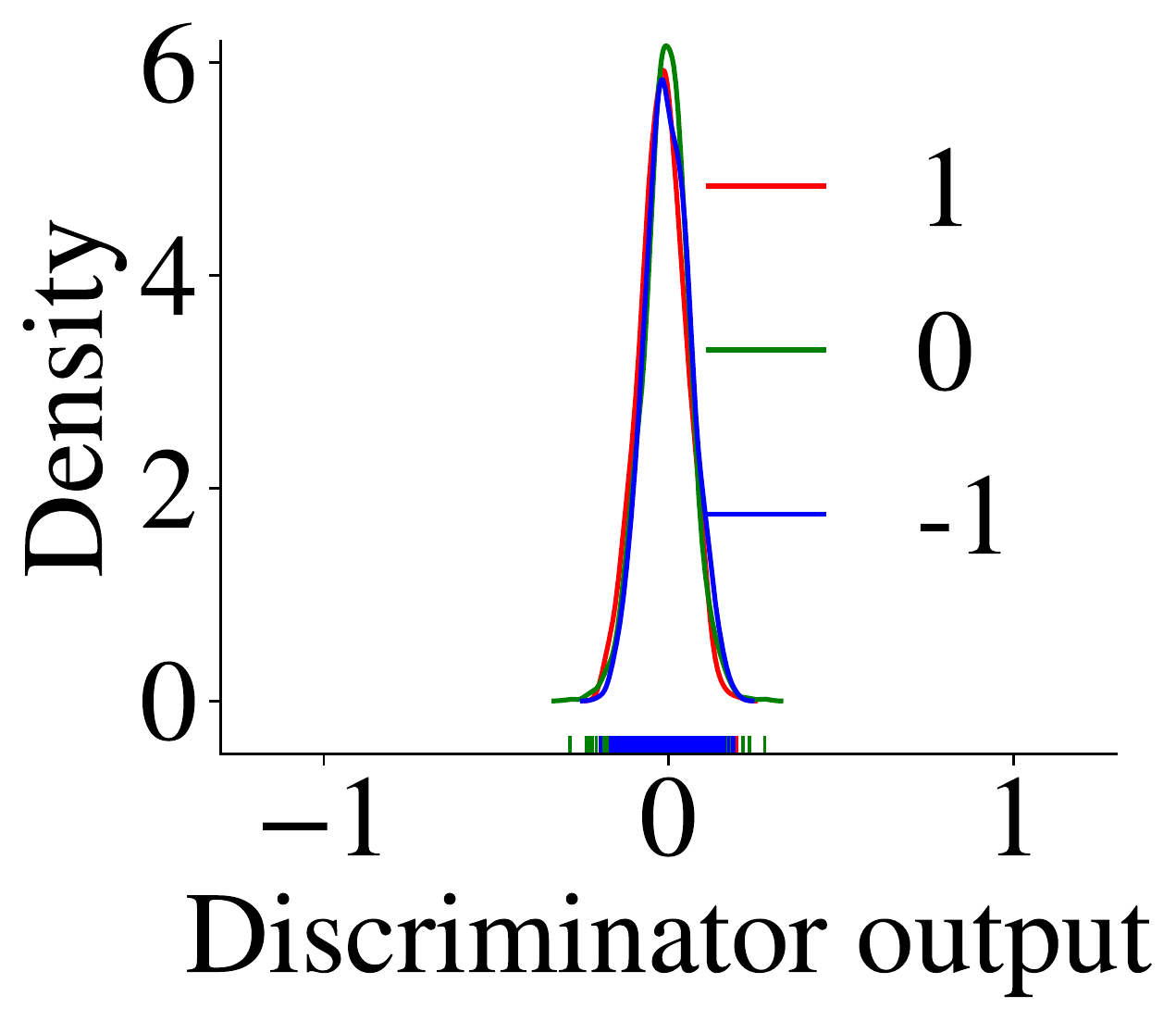}}
	\end{minipage}
	\hfill
	\begin{minipage}{0.32\linewidth}
	\centerline{\includegraphics[width=\textwidth]{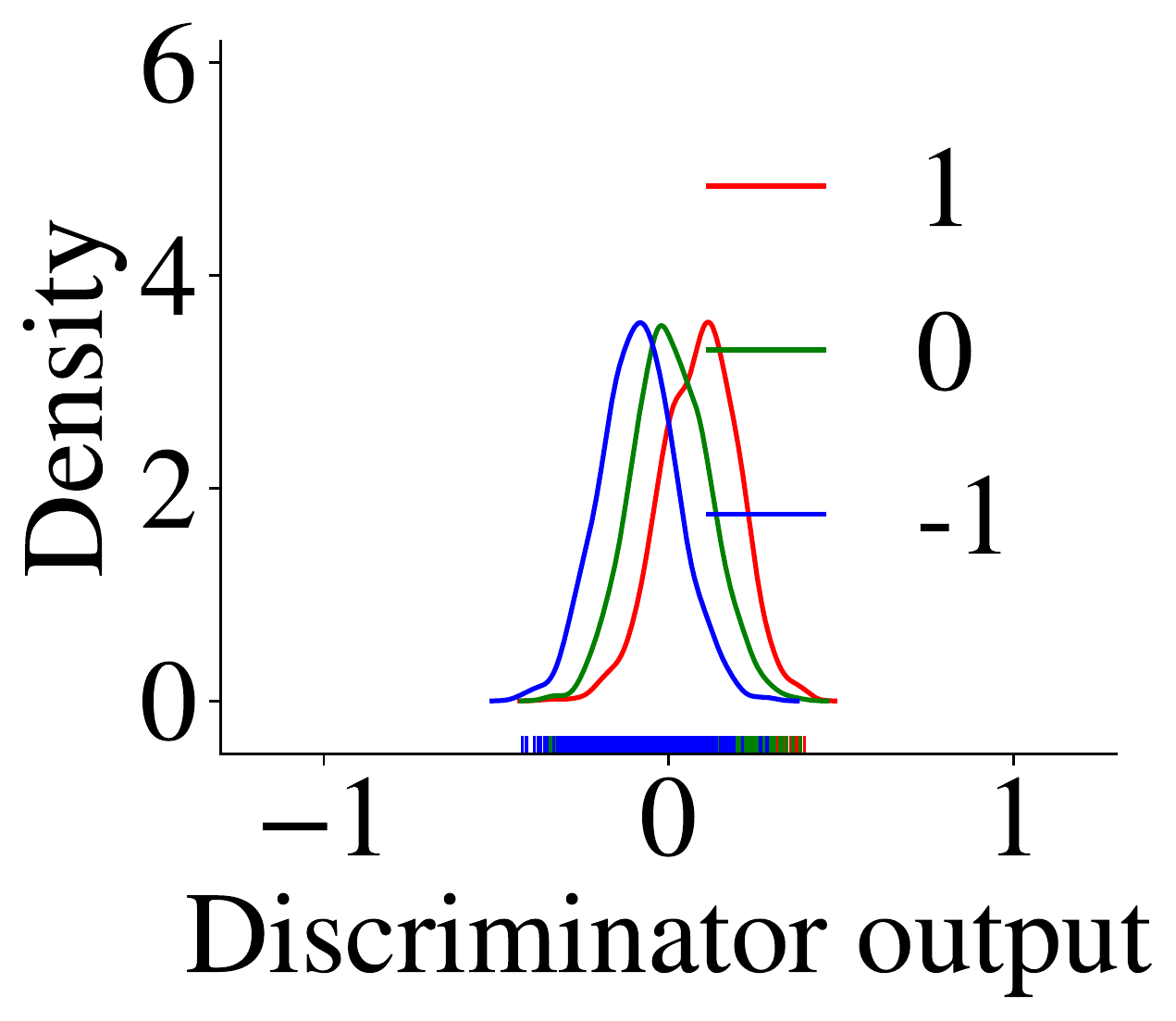}}
	\end{minipage}
	\hfill
	\begin{minipage}{0.32\linewidth}
 \centerline{\includegraphics[width=\textwidth]{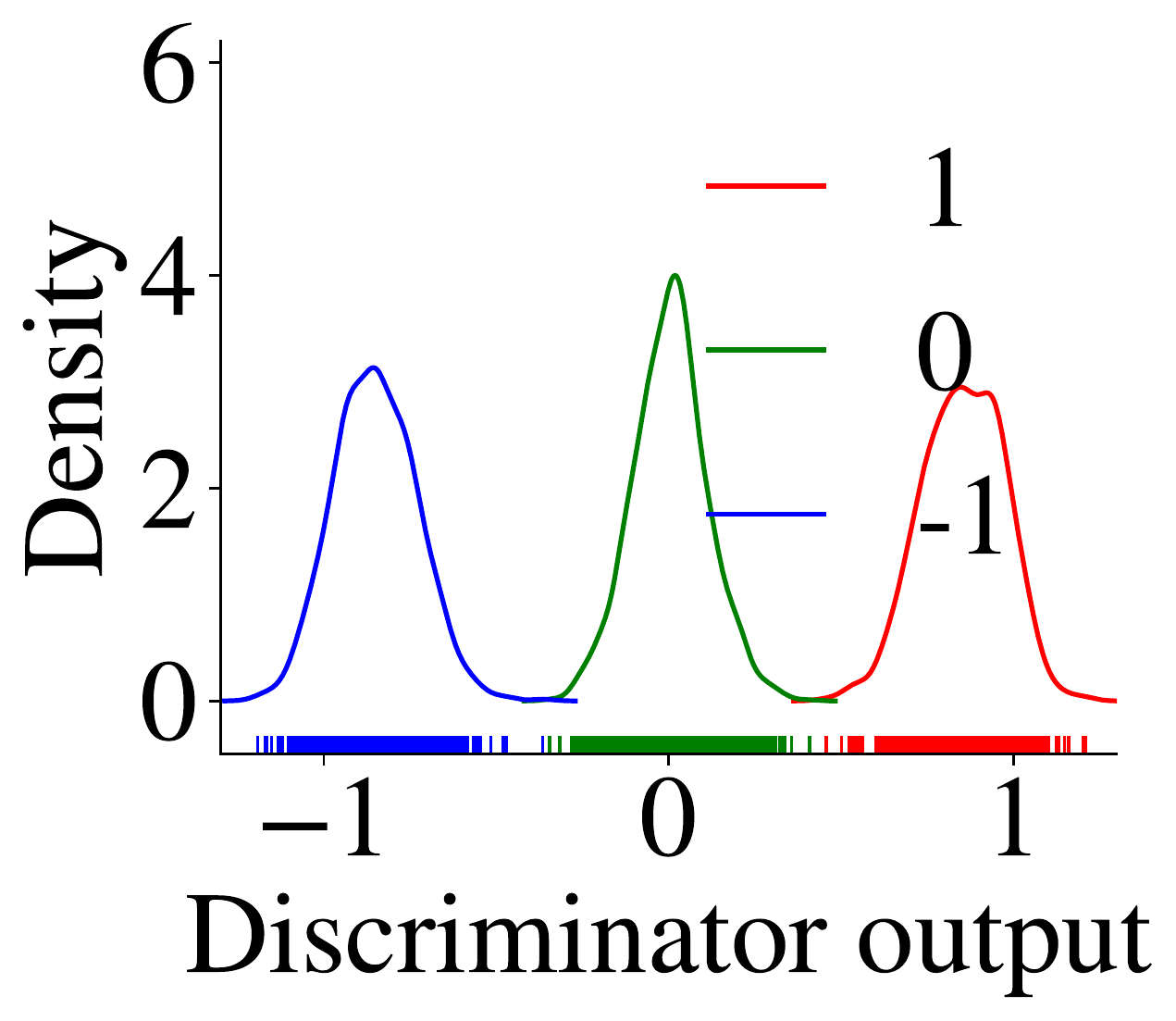}}
	\end{minipage}	
	\\
	\begin{minipage}{0.32\linewidth}
	\centerline{\includegraphics[width=\textwidth]{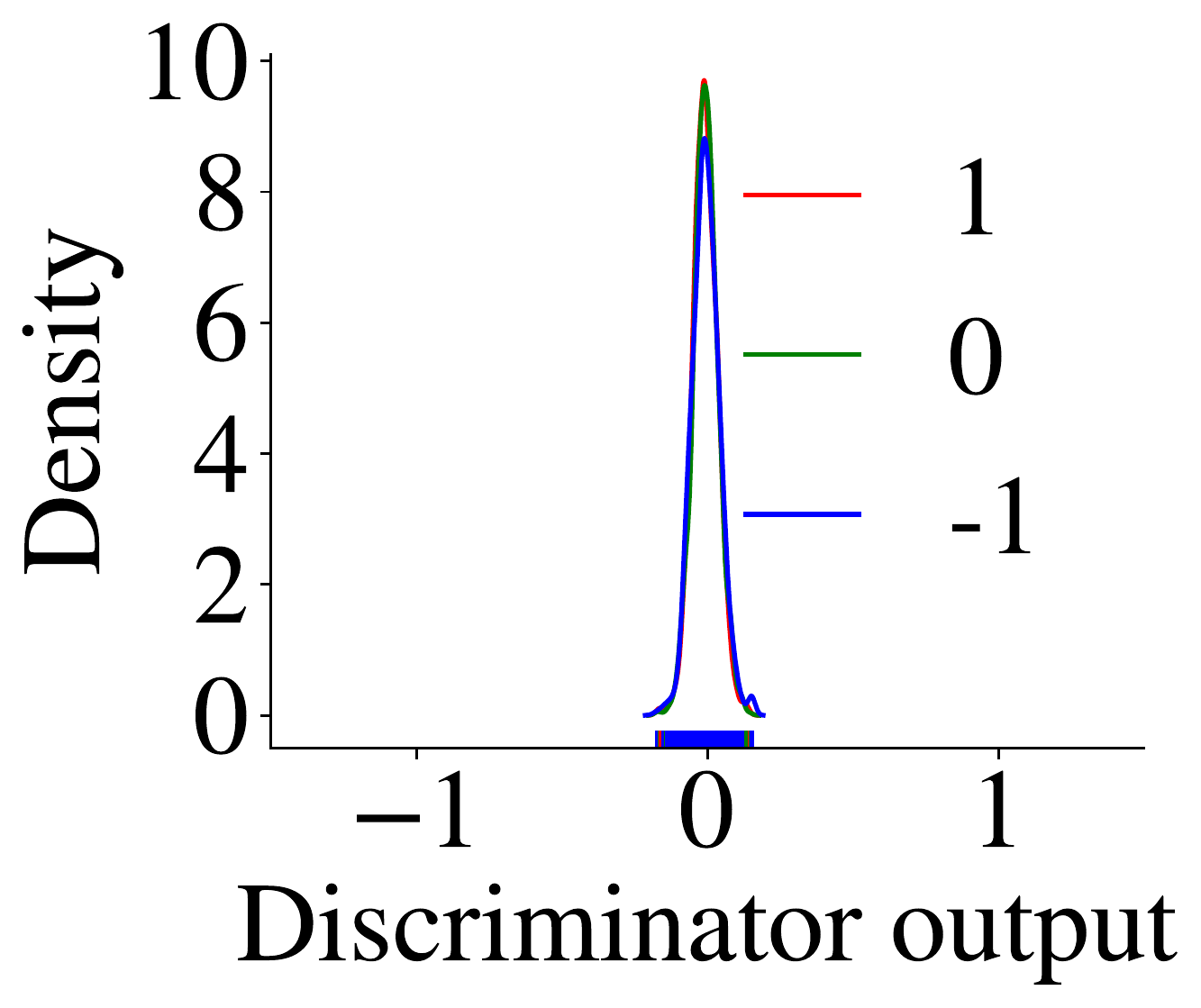}}
	\centerline{Iteration 0}
	\end{minipage}
	\hfill
	\begin{minipage}{0.32\linewidth}
	\centerline{\includegraphics[width=\textwidth]{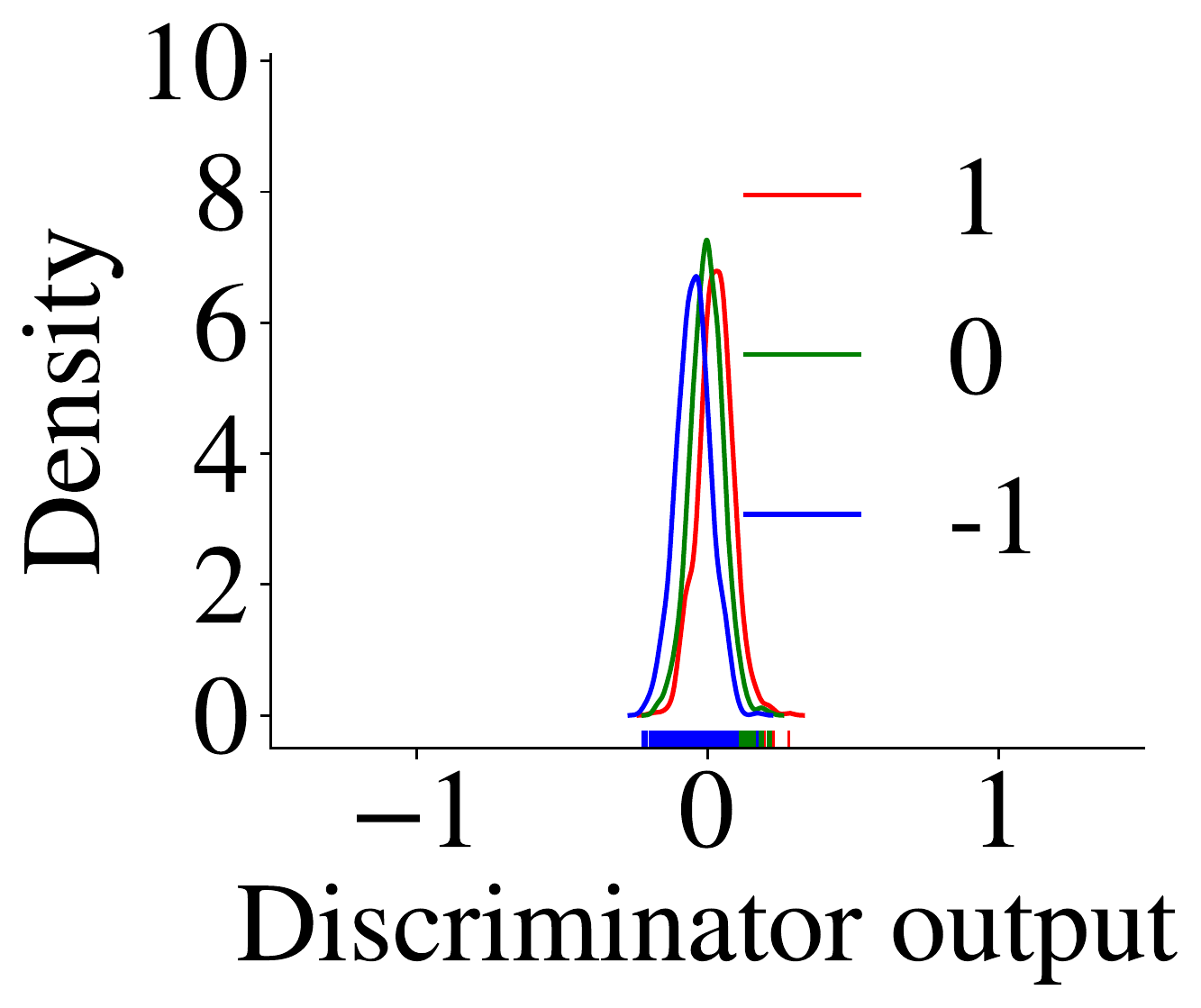}}
	\centerline{Iteration 1K}
	\end{minipage}
	\hfill
	\begin{minipage}{0.32\linewidth}
	\centerline{\includegraphics[width=\textwidth]{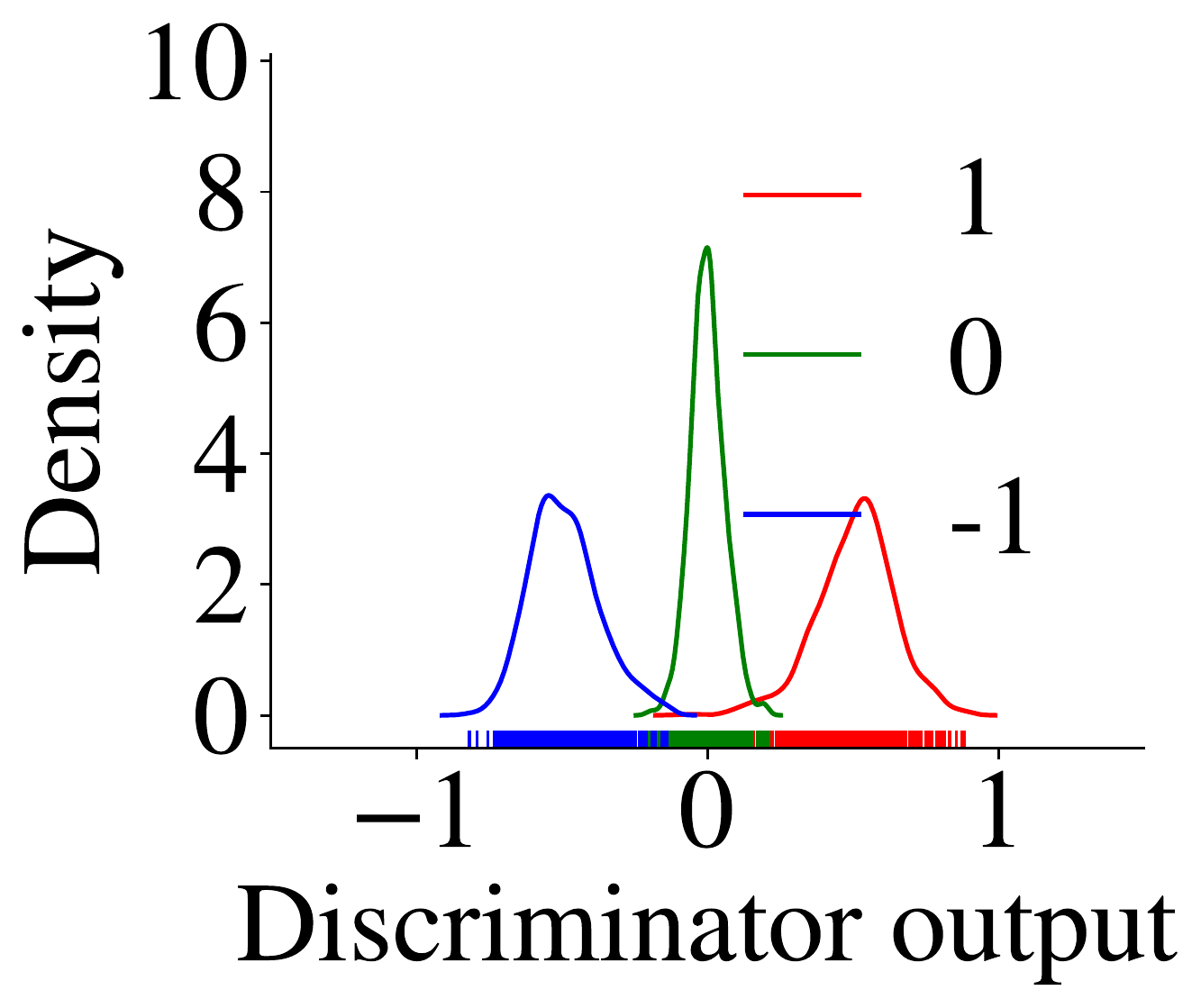}}
	\centerline{Iteration 99K}
	\end{minipage}		
  \\  
  \caption{\label{fig:density_pair}The density plot of the ranker's output for real pairs (first row) and generated pairs (second row) with different relative attributes ($+1/0/-1$), respectively, during the training process.}
\end{figure}

As our ranker can reveal the changes between image pairs, which is verified in the section~\ref{sect:ranker_meaning} and Fig.~\ref{fig:density_pair}, we use it to evaluate the subtle differences between the fine-grained synthetic image pairs generated by various baselines.

We generate the fine-grained pairs on the test dataset of CelebA-HQ w.r.t. the ``smile'' attribute.
Each trained model produces a series of synthetic images by taking as input a real image and different latent variables. 
The range of the latent variable is from -$1$ to $1$ with step $0.1$. Then the ranker, pre-trained by our TRIP,  is applied to evaluate the generated pairs and group them in terms of different conditioned latent variables for different models, respectively. In terms of each group, we calculate the mean and the standard deviation (std) for the outputs of the ranker (Fig.~\ref{fig:d_line}). 

Fig.~\ref{fig:d_line} shows that
(1) the ranking output of TRIP exhibits a linear trend with the lowest variance w.r.t. the latent variable. This demonstrates that TRIP can smoothly translate the input image to the desired image over the specific attribute along the latent variable. 
(2) The ranking output of RCGAN behaves like a $\tanh$ curve with a sudden change when the latent variable is around zero. It means that RCGAN cannot smoothly control the attribute strength for the input image. In addition, RCGAN has the largest variance on the ranking output due to the low quality of the generated images, which introduces noises to the ranker's prediction on the generated pairs. (3) RelGAN manifests a three-step like curve, which indicates a failure of fine-grained generation.  This is mainly because of its specific design of the interpolation loss. 
(4) FN presents a linear tendency like TRIP, which denotes that it can make a fine-grained control over the attribute. However, the mean of the ranking output for the generated pairs is relatively low in FN, since it fails to translate some input images into the desired output images. This is verified by its low translation accuracy (Fig.~\ref{fig:aas}), lower than $85\%$.  In addition, FN also exhibits a large variance of the ranking output due to the poor image quality.

\subsection{Ablation Study}
In Fig.~\ref{fig:ablation} and Table~\ref{tb:ablation}, we show an ablation study of our model. (1) Without $L_{rank}$, the generated images exhibit no change over the input image. The generator fails to learn the translation function, which is demonstrated by an extremely low translation accuracy (SSIM). (2) Without $L_{gan}$, the image quality degrades, achieving a high FID score. (3) Setting $\lambda=0$, i.e., without considering the adversarial ranking, the performance of facial image manipulation collapses, obtaining a low translation accuracy. (4) When optimizing with Eq.~\ref{eq:g_bin}, i.e., not linearing the ranking output for the generated pairs, the fine-grained control over the attributes fails, getting a high SSIM score. (5) With our TRIP, the generated images present desired changes consistent with the latent variable and possess good quality. 

\begin{table}[!ht]
\centering
\caption{\label{tb:ablation} Ablation study on each part of TRIP model.}
\renewcommand{\arraystretch}{1.2}
\setlength{\tabcolsep}{1.2mm}{	
\scalebox{1}{
\begin{tabular}{cccccc}
\toprule[1.3pt]
Model & $L_{rank}=0$ & $L_{gan}=0$ & $\lambda=0$ & CLS (3) & TRIP   \\
\midrule[1pt]
AAS              & 9.73         & 98.47       & 55.7        & 92.37   & 97.69  \\
SSIM             & 1.51E-05     & 0.0058      & 0.0011      & 0.0163  & 0.0030 \\
FID              & 29           & 78.08       & 8.55        & 11.33   & 10.19 \\
\toprule[1.3pt]
\end{tabular}}}
\vskip-0.2in
\end{table}

\subsection{Extension to Multiple Attributes}
We conduct fine-grained I2I translation with both ``smile'' and ``male'' on CelebA-HQ to show that TRIP can generalize well with multiple attributes. A two-dimension variable is used to control the change of both attributes, simultaneously.
\begin{figure}[!htb]
    \centering
    \includegraphics[width=0.95\linewidth]{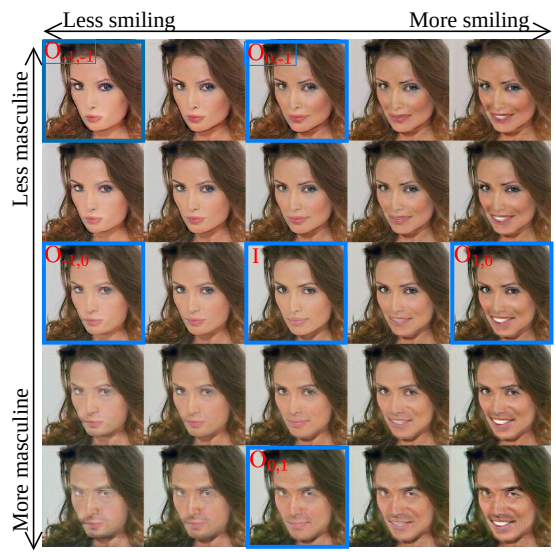}\vskip -0.1in
    \caption{\label{fig:multi_attr}Fine-grained I2I translation with ``smile'' and ``male'' attributes. The middle is the input image.}
\end{figure}

The generated images conditioning on different $\mathbf{v}$ are shown in Fig.~\ref{fig:multi_attr}. (1) TRIP can disentangle multiple attributes. When $\mathbf{v}=[-1, 0]/[1,0]$, the generated images $\text{O}_\text{-1,0}$/$\text{O}_\text{1,0}$ appear ``less smiling''/``more smiling'' with no change on the ``male'' attribute. When $\mathbf{v}=[0, -1]/[0, 1]$, the generated images $\text{O}_\text{0, -1}$/$\text{O}_\text{0,1}$ appear ``less masculine''/``more masculine'' with no change in the ``smiling'' attribute.  In addition, a fine-grained control over the strength of a single attribute is still practical.  
(2) TRIP can manipulates the subtle changes of multiple attributes simultaneously. For example, when conditioning $\mathbf{v}=[1,-1]$, the generated image $\text{O}_\text{1, -1}$ appear ``less smiling'' and ``more masculine''.

\subsection{Convergence of TRIP}
TRIP converges to an equilibrium when the generator produces the realistic image with desired changes over the input image regarding the target attribute, and the ranker makes reliable prediction for the difference between the translated images and the input image w.r.t the target attribute. 

\begin{figure}[!ht]
    \centering
    \includegraphics[width=.8\linewidth]{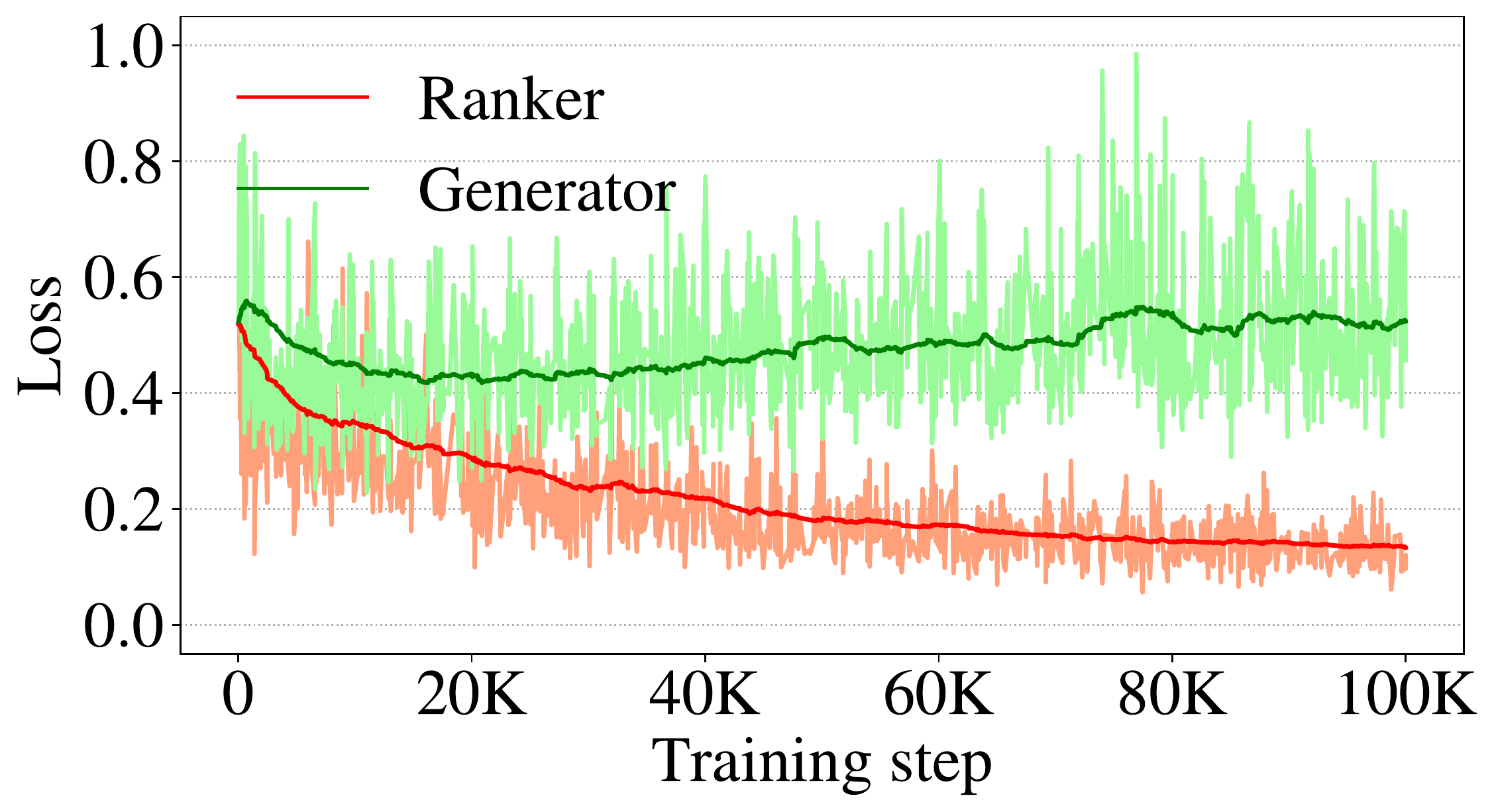}
    \caption{The curve of training loss. The ranker and the generator are trained against each other until convergence. $\text{Ranker loss}=L_{rank}^R+\lambda_g L_{gan}^R+\lambda_{gp}L_{gp}$. $\text{Generator loss}=L_{rank}^G+\lambda_g L_{gan}^G+\lambda_{gp}L_{cycle}$.}
    \label{fig:loss}
\end{figure}

To justify our claim, we plot the training curve of the ranker and the generator, respectively, as shown in Fig.~\ref{fig:loss}. It demonstrates that the ranker and the generator are trained against each other until convergence. 

Further, we plot the distribution of the ranker's prediction for real image pairs and generated image pairs with different relative attributes (RAs) ($+1/0/-1$) using the ranker in Fig.~\ref{fig:density_pair}. (1) At the beginning of the training, the ranker gives similar predictions for real image pairs with different RAs. The same observations can also be found on the generated image pairs. (2) After $100$ iterations, the ranker learns to give the desired prediction for different kinds of pair, i.e., $>0$ (averaged) for pairs with RA $(+1)$, $0$ (averaged) for pairs with RA $(0)$ and $<0$ (averaged) for pairs with RA $(-1)$. (3) After $9,900$ iterations, TRIP converges. In terms of the real image pairs, the ranker output $+1$ for the pairs with RA $(+1)$, $0$ for the pairs with RA $(0)$ and $-1$ for the pairs with RA $(-1)$ in the sense of average. \textbf{This verifies that our ranker can give precise ranking predictions for real image pairs.} In terms of the generated pairs, the ranker outputs $+0.5$ for the pairs with RA $(+1)$, $0$ for the pairs with RA $(0)$ and $-0.5$ for the pairs with RA $(-1)$ in the sense of average. \textbf{This is a convergence state due to rival preferences.} We take pairs with RA $(+1)$ as an example. The generated pairs with RA $(+1)$ are expected to be assigned $0$ when optimizing the ranker, and to be assigned $+1$ when optimizing the generator. Therefore, the convergence state should be around $0.5$. 

\section{Conclusion}
\label{sect:conclusion}
This paper proposes TRIP for high-quality fine-grained I2I translation. TRIP elegantly reconciles the goal for fine-grained translation and the goal for high-quality generation through adversarial ranking. It broadens the principle of adversarial training by extending the adversarial game on the binary classification to ranking. On the other hand, since the supervised pairwise ranking and the unsupervised generation target are incorporated into a single model function, TRIP can be deemed as a new form of semi-supervised GAN~\cite{odena2016semi}. The empirical experiments demonstrate that TRIP achieves state-of-art results in generating high-fidelity images that exhibit smooth changes exclusively w.r.t. the specified attributes according to the fine-grained score and the quality score. One interesting future direction is to extend our idea to semi-supervised ranking~\cite{DBLP:journals/csl/DuhK11} given that it has been theoretically proven that semi-supervised GAN can improve the performance of semi-supervised classification~\cite{DBLP:conf/nips/DaiYYCS17}.

\bibliographystyle{IEEEtran}
\bibliography{TRIP_bare_jrnl_compsoc}

\begin{thebibliography}{10}
\providecommand{\url}[1]{#1}
\csname url@samestyle\endcsname
\providecommand{\newblock}{\relax}
\providecommand{\bibinfo}[2]{#2}
\providecommand{\BIBentrySTDinterwordspacing}{\spaceskip=0pt\relax}
\providecommand{\BIBentryALTinterwordstretchfactor}{4}
\providecommand{\BIBentryALTinterwordspacing}{\spaceskip=\fontdimen2\font plus
\BIBentryALTinterwordstretchfactor\fontdimen3\font minus
  \fontdimen4\font\relax}
\providecommand{\BIBforeignlanguage}[2]{{%
\expandafter\ifx\csname l@#1\endcsname\relax
\typeout{** WARNING: IEEEtran.bst: No hyphenation pattern has been}%
\typeout{** loaded for the language `#1'. Using the pattern for}%
\typeout{** the default language instead.}%
\else
\language=\csname l@#1\endcsname
\fi
#2}}
\providecommand{\BIBdecl}{\relax}
\BIBdecl

\bibitem{isola2017image}
P.~Isola, J.-Y. Zhu, T.~Zhou, and A.~A. Efros, ``Image-to-image translation
  with conditional adversarial networks,'' in \emph{Proceedings of the IEEE
  conference on computer vision and pattern recognition}, 2017, pp. 1125--1134.

\bibitem{zhu2017unpaired}
J.-Y. Zhu, T.~Park, P.~Isola, and A.~A. Efros, ``Unpaired image-to-image
  translation using cycle-consistent adversarial networks,'' in
  \emph{Proceedings of the IEEE international conference on computer vision},
  2017, pp. 2223--2232.

\bibitem{kim2017learning}
T.~Kim, M.~Cha, H.~Kim, J.~K. Lee, and J.~Kim, ``Learning to discover
  cross-domain relations with generative adversarial networks,'' in
  \emph{Proceedings of the 34th International Conference on Machine
  Learning-Volume 70}.\hskip 1em plus 0.5em minus 0.4em\relax JMLR. org, 2017,
  pp. 1857--1865.

\bibitem{9367012}
L.~Dai and J.~Tang, ``iflowgan: An invertible flow-based generative adversarial
  network for unsupervised image-to-image translation,'' \emph{IEEE
  Transactions on Pattern Analysis and Machine Intelligence}, pp. 1--1, 2021.

\bibitem{lample2017fader}
G.~Lample, N.~Zeghidour, N.~Usunier, A.~Bordes, L.~Denoyer, and M.~Ranzato,
  ``Fader networks: Manipulating images by sliding attributes,'' in
  \emph{Advances in Neural Information Processing Systems}, 2017, pp.
  5967--5976.

\bibitem{he2019attgan}
Z.~He, W.~Zuo, M.~Kan, S.~Shan, and X.~Chen, ``Attgan: Facial attribute editing
  by only changing what you want,'' \emph{IEEE Transactions on Image
  Processing}, vol.~28, no.~11, pp. 5464--5478, 2019.

\bibitem{liu2018unified}
A.~H. Liu, Y.-C. Liu, Y.-Y. Yeh, and Y.-C.~F. Wang, ``A unified feature
  disentangler for multi-domain image translation and manipulation,'' in
  \emph{Advances in neural information processing systems}, 2018, pp.
  2590--2599.

\bibitem{DBLP:conf/bmvc/SaquilKH18}
Y.~Saquil, K.~I. Kim, and P.~M. Hall, ``Ranking cgans: Subjective control over
  semantic image attributes,'' in \emph{British Machine Vision Conference 2018,
  {BMVC} 2018, Newcastle, UK, September 3-6, 2018}.\hskip 1em plus 0.5em minus
  0.4em\relax {BMVA} Press, 2018, p. 131.

\bibitem{li2019latent}
X.~Li, C.~Lin, C.~Wang, and F.~Guerin, ``Latent space factorisation and
  manipulation via matrix subspace projection,'' \emph{International Conference
  on Machine Learning}, 2020.

\bibitem{ding2020guided}
Z.~Ding, Y.~Xu, W.~Xu, G.~Parmar, Y.~Yang, M.~Welling, and Z.~Tu, ``Guided
  variational autoencoder for disentanglement learning,'' in \emph{Proceedings
  of the IEEE/CVF Conference on Computer Vision and Pattern Recognition}, 2020,
  pp. 7920--7929.

\bibitem{liu2019conditional}
R.~Liu, Y.~Liu, X.~Gong, X.~Wang, and H.~Li, ``Conditional adversarial
  generative flow for controllable image synthesis,'' in \emph{Proceedings of
  the IEEE/CVF Conference on Computer Vision and Pattern Recognition}, 2019,
  pp. 7992--8001.

\bibitem{8886528}
J.~Lin, Z.~Chen, Y.~Xia, S.~Liu, T.~Qin, and J.~Luo, ``Exploring explicit
  domain supervision for latent space disentanglement in unpaired
  image-to-image translation,'' \emph{IEEE Transactions on Pattern Analysis and
  Machine Intelligence}, vol.~43, no.~4, pp. 1254--1266, 2021.

\bibitem{wu2019relgan}
P.-W. Wu, Y.-J. Lin, C.-H. Chang, E.~Y. Chang, and S.-W. Liao, ``Relgan:
  Multi-domain image-to-image translation via relative attributes,'' in
  \emph{Proceedings of the IEEE International Conference on Computer Vision},
  2019, pp. 5914--5922.

\bibitem{parikh2011relative}
D.~Parikh and K.~Grauman, ``Relative attributes,'' in \emph{2011 International
  Conference on Computer Vision}.\hskip 1em plus 0.5em minus 0.4em\relax IEEE,
  2011, pp. 503--510.

\bibitem{goodfellow2014generative}
I.~Goodfellow, J.~Pouget-Abadie, M.~Mirza, B.~Xu, D.~Warde-Farley, S.~Ozair,
  A.~Courville, and Y.~Bengio, ``Generative adversarial nets,'' in
  \emph{Advances in neural information processing systems}, 2014, pp.
  2672--2680.

\bibitem{berthelot2018understanding}
D.~Berthelot, C.~Raffel, A.~Roy, and I.~Goodfellow, ``Understanding and
  improving interpolation in autoencoders via an adversarial regularizer,'' in
  \emph{International Conference on Learning Representations}, 2018.

\bibitem{kondo2019flow}
R.~Kondo, K.~Kawano, S.~Koide, and T.~Kutsuna, ``Flow-based image-to-image
  translation with feature disentanglement,'' in \emph{Advances in Neural
  Information Processing Systems}, 2019, pp. 4168--4178.

\bibitem{deng2020disentangled}
Y.~Deng, J.~Yang, D.~Chen, F.~Wen, and X.~Tong, ``Disentangled and controllable
  face image generation via 3d imitative-contrastive learning,'' in
  \emph{Proceedings of the IEEE/CVF Conference on Computer Vision and Pattern
  Recognition}, 2020, pp. 5154--5163.

\bibitem{alharbi2020disentangled}
Y.~Alharbi and P.~Wonka, ``Disentangled image generation through structured
  noise injection,'' in \emph{Proceedings of the IEEE/CVF Conference on
  Computer Vision and Pattern Recognition}, 2020, pp. 5134--5142.

\bibitem{cao2006adapting}
Y.~Cao, J.~Xu, T.-Y. Liu, H.~Li, Y.~Huang, and H.-W. Hon, ``Adapting ranking
  svm to document retrieval,'' in \emph{Proceedings of the 29th annual
  international ACM SIGIR conference on Research and development in information
  retrieval}, 2006, pp. 186--193.

\bibitem{odena2016semi}
A.~Odena, ``Semi-supervised learning with generative adversarial networks,''
  \emph{Workshop on Data-Efficient Machine Learning (ICML)}, 2016.

\bibitem{chapelle2008analysis}
O.~Chapelle, A.~Agarwal, F.~H. Sinz, and B.~Sch{\"o}lkopf, ``An analysis of
  inference with the universum,'' in \emph{Advances in neural information
  processing systems}, 2008, pp. 1369--1376.

\bibitem{10.1145/1390334.1390382}
K.~Zhou, G.-R. Xue, H.~Zha, and Y.~Yu, ``Learning to rank with ties,'' in
  \emph{Proceedings of the 31st Annual International ACM SIGIR Conference on
  Research and Development in Information Retrieval}, 2008, p. 275–282.

\bibitem{DBLP:conf/iclr/KarrasALL18}
T.~Karras, T.~Aila, S.~Laine, and J.~Lehtinen, ``Progressive growing of gans
  for improved quality, stability, and variation,'' in \emph{6th International
  Conference on Learning Representations, {ICLR} 2018, Vancouver, BC, Canada,
  April 30 - May 3, 2018, Conference Track Proceedings}.\hskip 1em plus 0.5em
  minus 0.4em\relax OpenReview.net, 2018.

\bibitem{liu2015faceattributes}
Z.~Liu, P.~Luo, X.~Wang, and X.~Tang, ``Deep learning face attributes in the
  wild,'' in \emph{Proceedings of International Conference on Computer Vision
  (ICCV)}, December 2015.

\bibitem{finegrained}
A.~Yu and K.~Grauman, ``Fine-grained visual comparisons with local learning,''
  in \emph{IEEE Conference on Computer Vision and Pattern Recognition}, Jun
  2014.

\bibitem{DBLP:conf/iclr/ZhangCDL18}
H.~Zhang, M.~Ciss{\'{e}}, Y.~N. Dauphin, and D.~Lopez{-}Paz, ``mixup: Beyond
  empirical risk minimization,'' in \emph{6th International Conference on
  Learning Representations}.\hskip 1em plus 0.5em minus 0.4em\relax
  OpenReview.net, 2018.

\bibitem{wang2004image}
Z.~Wang, A.~C. Bovik, H.~R. Sheikh, and E.~P. Simoncelli, ``Image quality
  assessment: from error visibility to structural similarity,'' \emph{IEEE
  transactions on image processing}, vol.~13, no.~4, pp. 600--612, 2004.

\bibitem{he2016deep}
K.~He, X.~Zhang, S.~Ren, and J.~Sun, ``Deep residual learning for image
  recognition,'' in \emph{Proceedings of the IEEE conference on computer vision
  and pattern recognition}, 2016, pp. 770--778.

\bibitem{DBLP:journals/csl/DuhK11}
K.~Duh and K.~Kirchhoff, ``Semi-supervised ranking for document retrieval,''
  \emph{Comput. Speech Lang.}, vol.~25, no.~2, pp. 261--281, 2011.

\bibitem{DBLP:conf/nips/DaiYYCS17}
Z.~Dai, Z.~Yang, F.~Yang, W.~W. Cohen, and R.~Salakhutdinov, ``Good
  semi-supervised learning that requires a bad {GAN},'' in \emph{Advances in
  Neural Information Processing Systems}, 2017, pp. 6510--6520.

\end{thebibliography}

\onecolumn
\newpage
\appendices
\section{Comparison of TRIP with RCGAN and RelGAN}
\begin{figure*}[!htb]
\centering
	\begin{minipage}{\linewidth}
    \centerline{\includegraphics[width=.9\textwidth]{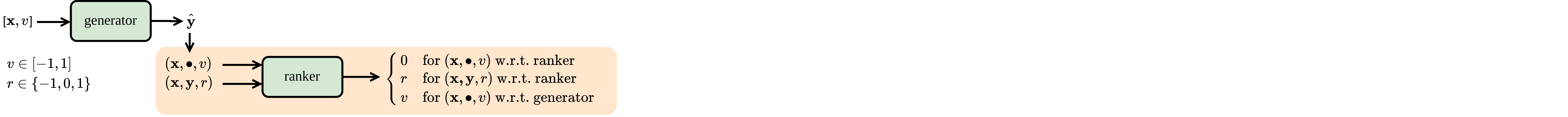}}
	\centerline{(a) TRIP}
	\end{minipage}\\ \vskip0.2in
	\begin{minipage}{\linewidth}
	\centerline{\includegraphics[width=.9\textwidth]{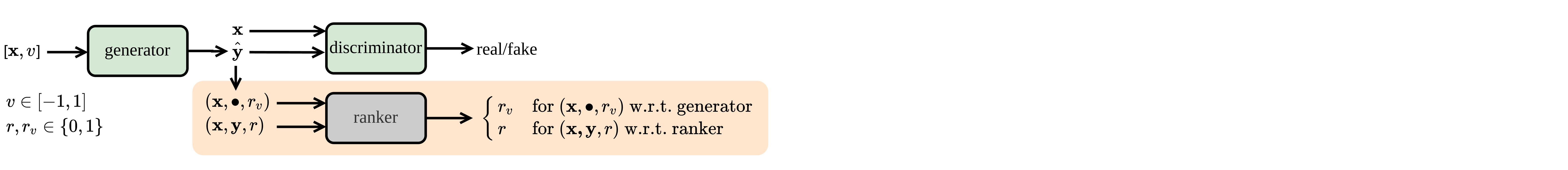}}
	\centerline{(b) RCGAN}
	\end{minipage}
	\\ \vskip0.1in
	\begin{minipage}{\linewidth}
 \centerline{\includegraphics[width=\textwidth]{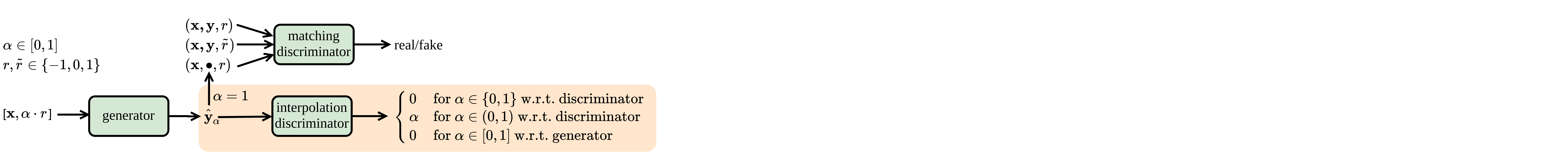}}
 	\centerline{(c) RelGAN}
	\end{minipage}
	 \caption{\label{fig:trip_rcgan_relgan_arch} The architectures of TRIP, RCGAN and RelGAN. The orange shade denotes the key differences between TRIP and RCGAN and RelGAN. \textbf{TRIP} re-designs the framework for the adversarial game while the original game is adversarial binary classification proposed in GAN). TRIP designs adversarial ranking by rival preferences, which is the opposite goals (0/$v$) between the ranker and the generator w.r.t. the generated pairs $(\mathbf{x}, \hat{\mathbf{y}})$. \textbf{RCGAN} inherits adversarial classification from GAN for distribution matching between real data and generated data. To distill discrepancy from relative attributes (RAs) for fine-grained generation, an additional ranker is introduced beyond vanilla GAN framework. Note that the generated pairs $(\mathbf{x}, \hat{\mathbf{y}})$ are not used to train the ranker, denoted in grey shade. \textbf{RelGAN} inherits adversarial binary classification from GAN but for joint distribution matching between real triplets $(\mathbf{x}, \mathbf{y}, r)$ and fake/wrong triplets $(\mathbf{x}, \hat{\mathbf{y}}, r)$/$(\mathbf{x}, \mathbf{y}, \tilde{r})$. The interpolation discriminator is to enforce the good quality of images with interpolated RAs.}
\end{figure*}

\newpage
\section{Network Architectures}
\begin{table*}[!htb]
\centering
\caption{\label{tb:D_structure} Ranker network architecture. LReLU is Leaky ReLU with a negative slop of $0.01$. $K$ is the number of attributes. $N_f$ is the number of filters. $S_f$ is the filter size. $S_s$ is the stride size. $S_p$ is the padding size.}
\renewcommand{\arraystretch}{1.1}
\setlength{\tabcolsep}{1.1mm}{	
\scalebox{1}{
\begin{tabular}{ccc}
\toprule[1.3pt]
Component            & Input $\rightarrow$ Output Shape                                                                                                           & Layer Information                                                   \\ \hline
Feature Layer        & $2\times(h , w, 3) \rightarrow 2\times\left(\frac{h}{2}, \frac{w}{2}, 64\right)$                                                           & Conv-($N_f$=64,$S_f$=4,$S_s$=2,$S_p$=1),LReLU                       \\
                     & $2\times\left(\frac{h}{2}, \frac{w}{2}, 64\right)\rightarrow 2\times\left(\frac{h}{4}, \frac{w}{4}, 128\right)$                            & Conv-($N_f$=128,$S_f$=4,$S_s$=2,$S_p$=1),LReLU                      \\
\multicolumn{1}{l}{} & \multicolumn{1}{l}{$2\times\left(\frac{h}{4}, \frac{w}{4}, 128\right) \rightarrow 2\times\left(\frac{h}{8}, \frac{w}{8}, 256\right)$}      & \multicolumn{1}{l}{Conv-($N_f$=256,$S_f$=4,$S_s$=2,$S_p$=1),LReLU}  \\
\multicolumn{1}{l}{} & \multicolumn{1}{l}{$2\times\left(\frac{h}{8}, \frac{w}{8}, 256\right) \rightarrow 2\times\left(\frac{h}{16}, \frac{w}{16}, 512\right)$}    & \multicolumn{1}{l}{Conv-($N_f$=512,$S_f$=4,$S_s$=2,$S_p$=1),LReLU}  \\
\multicolumn{1}{l}{} & \multicolumn{1}{l}{$2\times\left(\frac{h}{16}, \frac{w}{16}, 512\right) \rightarrow 2\times\left(\frac{h}{32}, \frac{w}{32}, 1024\right)$} & \multicolumn{1}{l}{Conv-($N_f$=1024,$S_f$=4,$S_s$=2,$S_p$=1),LReLU} \\
                     & $2\times\left(\frac{h}{32}, \frac{w}{32}, 1024\right) \rightarrow 2\times\left(\frac{h}{64}, \frac{w}{64}, 2048\right)$                    & Conv-($N_f$=2048,$S_f$=4,$S_s$=2,$S_p$=1),LReLU                     \\ \hline
Rank Layer           & $2\times\left(\frac{h}{64}, \frac{w}{64}, 2048\right)\rightarrow \left(\frac{h}{64}, \frac{w}{64}, 2048\right)$                            & Subtract                                                            \\
                     & $\left(\frac{h}{64}, \frac{w}{64}, 2048\right)\rightarrow\left(\frac{h}{64}, \frac{w}{64}, 2048\right)$                                    & Conv-($N_f$=1,$S_f$=1,$S_s$=1,$S_p$=1),LReLU                        \\
                     & $\left(\frac{h}{64}, \frac{w}{64}, 2048\right) \rightarrow\left(\frac{h}{64} \times \frac{w}{64} \times 2048\right)$                       & Flatten                                                             \\
                     & $\left(\frac{h}{64} \times \frac{w}{64} \times 2048\right) \rightarrow (K,)$                                                               & Dense                                                               \\ \hline
GAN Layer            & $\left(\frac{h}{64}, \frac{w}{64}, 2048\right) \rightarrow\left(\frac{h}{64} \times \frac{w}{64} \times 2048\right)$                       & Flatten                                                             \\
                     & $\left(\frac{h}{64} \times \frac{w}{64} \times 2048\right)\rightarrow (1,)$                                                                & Dense                                                               \\
\toprule[1.3pt]
\end{tabular}}}
\end{table*}

\begin{table*}[!htb]
\centering
\caption{\label{tb:G_structure} Generator network architecture. We use switchable normalization, denoted as SN, in all layers except the last output layer. $N_f$ is the number of filters. $S_f$ is the filter size. $S_s$ is the stride size. $S_p$ is the padding size.}
\renewcommand{\arraystretch}{1.1}
\setlength{\tabcolsep}{1.1mm}{	
\scalebox{1}{
\begin{tabular}{ccc}
\toprule[1.3pt]
Component       & Input $\rightarrow$ Output Shape                                                                   & Layer Information                                                \\ \hline
Down-sampling   & $\left(h, w, 3+K\right) \rightarrow(h, w, 64)$                                                     & Conv-($N_f$=64,$S_f$=7,$S_s$=1,$S_p$=3),SN,ReLU                  \\
                & $(h, w, 64) \rightarrow\left(\frac{h}{2}, \frac{w}{2}, 128\right)$                                 & Conv-($N_f$=128,$S_f$=4,$S_s$=2,$S_p$=1),SN,ReLU                 \\
                & $\left(\frac{h}{2}, \frac{w}{2}, 128\right) \rightarrow\left(\frac{h}{4}, \frac{w}{4}, 256\right)$ & Conv-($N_f$=256,$S_f$=4,$S_s$=2,$S_p$=1),SN,ReLU                 \\ \hline
Residual Blocks & $\left(\frac{h}{4}, \frac{w}{4}, 256\right) \rightarrow\left(\frac{h}{4}, \frac{w}{4}, 256\right)$ & Residual Block: Conv-($N_f$=256,$S_f$=3,$S_s$=1,$S_p$=1),SN,ReLU \\
                & $\left(\frac{h}{4}, \frac{w}{4}, 256\right) \rightarrow\left(\frac{h}{4}, \frac{w}{4}, 256\right)$ & Residual Block: Conv-($N_f$=256,$S_f$=3,$S_s$=1,$S_p$=1),SN,ReLU \\
                & $\left(\frac{h}{4}, \frac{w}{4}, 256\right) \rightarrow\left(\frac{h}{4}, \frac{w}{4}, 256\right)$ & Residual Block: Conv-($N_f$=256,$S_f$=3,$S_s$=1,$S_p$=1),SN,ReLU \\
                & $\left(\frac{h}{4}, \frac{w}{4}, 256\right) \rightarrow\left(\frac{h}{4}, \frac{w}{4}, 256\right)$ & Residual Block: Conv-($N_f$=256,$S_f$=3,$S_s$=1,$S_p$=1),SN,ReLU \\
                & $\left(\frac{h}{4}, \frac{w}{4}, 256\right) \rightarrow\left(\frac{h}{4}, \frac{w}{4}, 256\right)$ & Residual Block: Conv-($N_f$=256,$S_f$=3,$S_s$=1,$S_p$=1),SN,ReLU \\
                & $\left(\frac{h}{4}, \frac{w}{4}, 256\right) \rightarrow\left(\frac{h}{4}, \frac{w}{4}, 256\right)$ & Residual Block: Conv-($N_f$=256,$S_f$=3,$S_s$=1,$S_p$=1),SN,ReLU \\ \hline
Up-sampling     & $\left(\frac{h}{4}, \frac{w}{4}, 256\right) \rightarrow\left(\frac{h}{2}, \frac{w}{2}, 128\right)$ & Conv-($N_f$=128,$S_f$=4,$S_s$=2,$S_p$=1),SN,ReLU                 \\
                & $\left(\frac{h}{2}, \frac{w}{2}, 128\right) \rightarrow(h, w, 64)$                                 & Conv-($N_f$=64,$S_f$=4,$S_s$=2,$S_p$=1),SN,ReLU                  \\
                & $(h, w, 64) \rightarrow(h, w, 3)$                                                                  & Conv-($N_f$=3,$S_f$=7,$S_s$=1,$S_p$=3),Tanh                      \\
\toprule[1.3pt]
\end{tabular}}}
\end{table*}

\section{More Experimental Results}
\begin{landscape}
\begin{figure*}[!htb]
\vskip-0.3in
\centering
    \begin{minipage}{0.9\linewidth}
    \centerline{\includegraphics[width=1\textwidth]{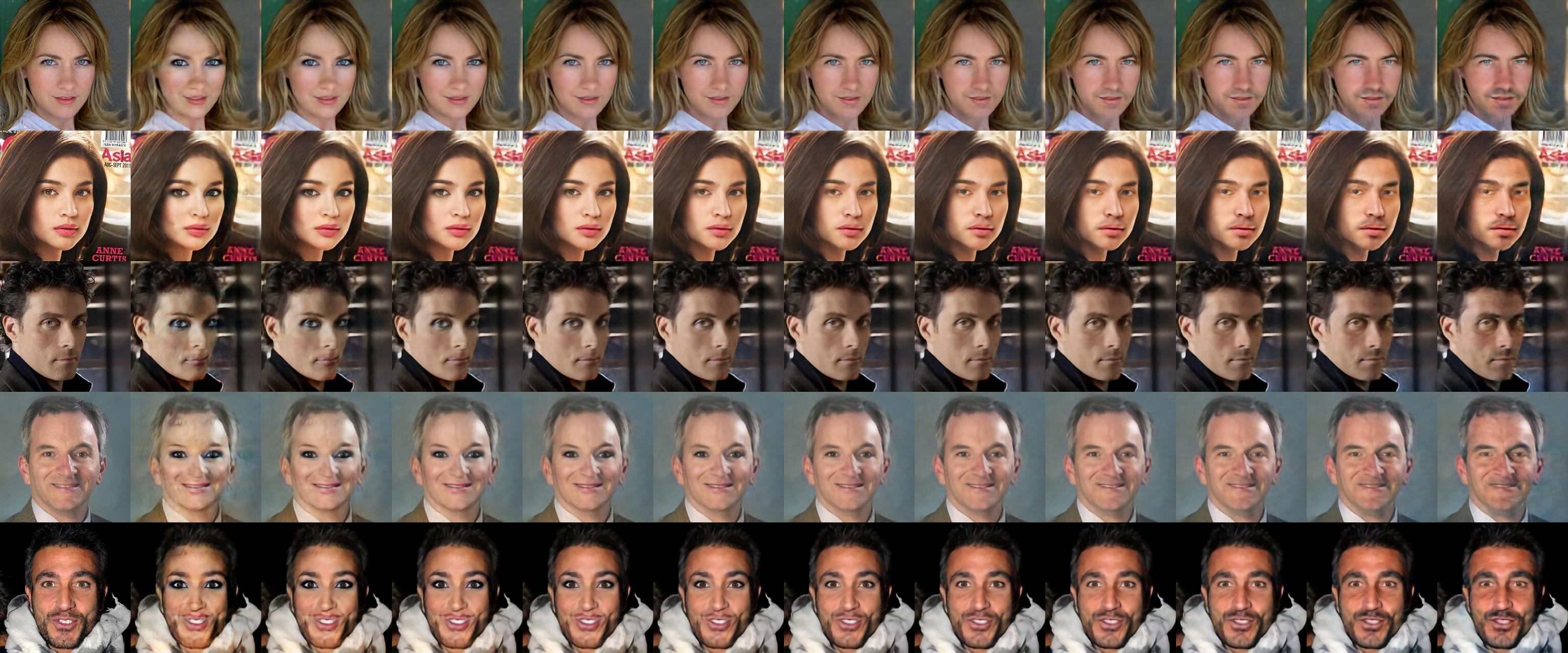}}
	\centerline{FN}
	\end{minipage}
	\\
	\begin{minipage}{0.9\linewidth}
	\centerline{\includegraphics[width=1\textwidth]{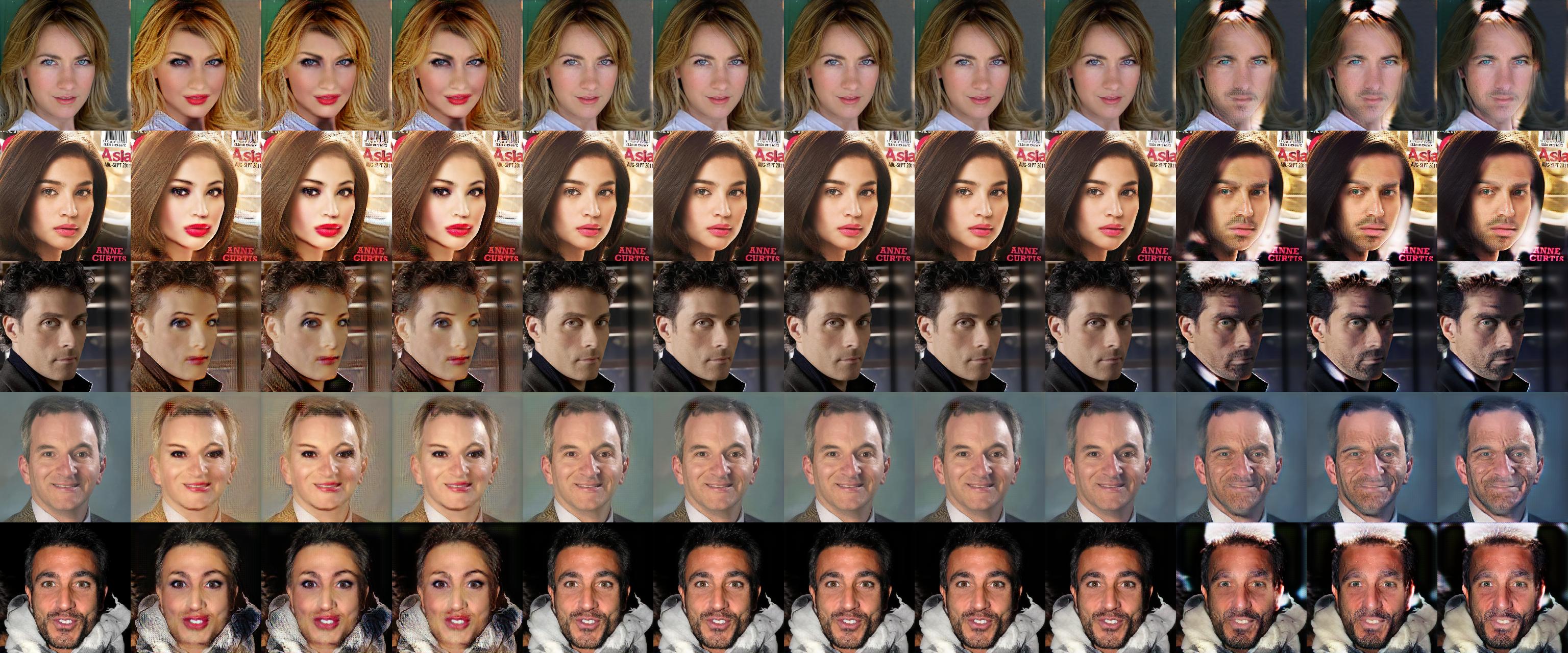}}
	\centerline{RelGAN}
	\end{minipage}
	\vskip-0.1in
\caption{\label{fig:celebahq_male} The fine-grained facial attribute (``male'') translation on CelebA-HQ dataset by FN and RelGAN. The first column is the input image. The other columns are the generated images conditioned on the latent variable from $-1$ to $1$ with step 0.2. The presented results are randomly sampled from the test set.}
\end{figure*}
\end{landscape}

\begin{landscape}
\begin{figure*}[!htb]
\vskip-0.3in
\centering
	\begin{minipage}{0.9\linewidth}
 \centerline{\includegraphics[width=\textwidth]{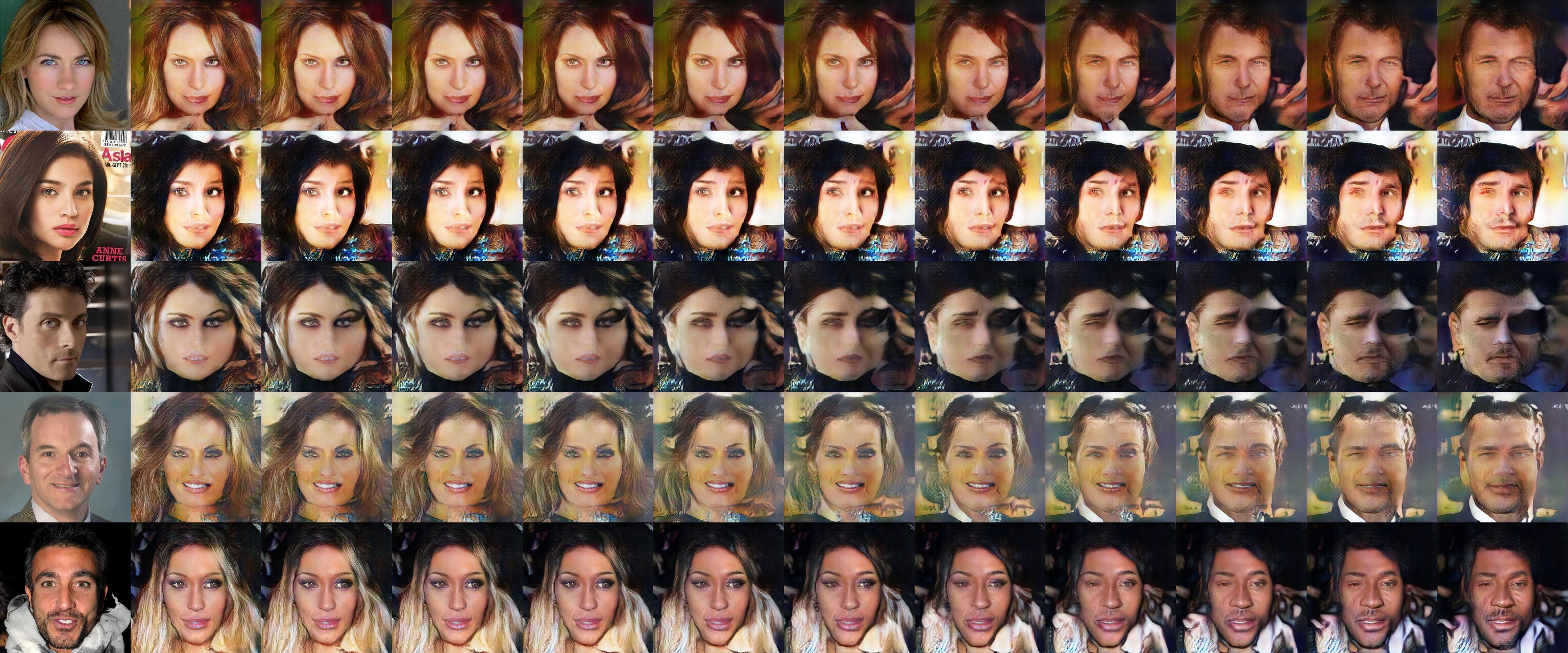}}
 	\centerline{RCGAN}
	\end{minipage}	
	\\
	\begin{minipage}{0.9\linewidth}
 \centerline{\includegraphics[width=\textwidth]{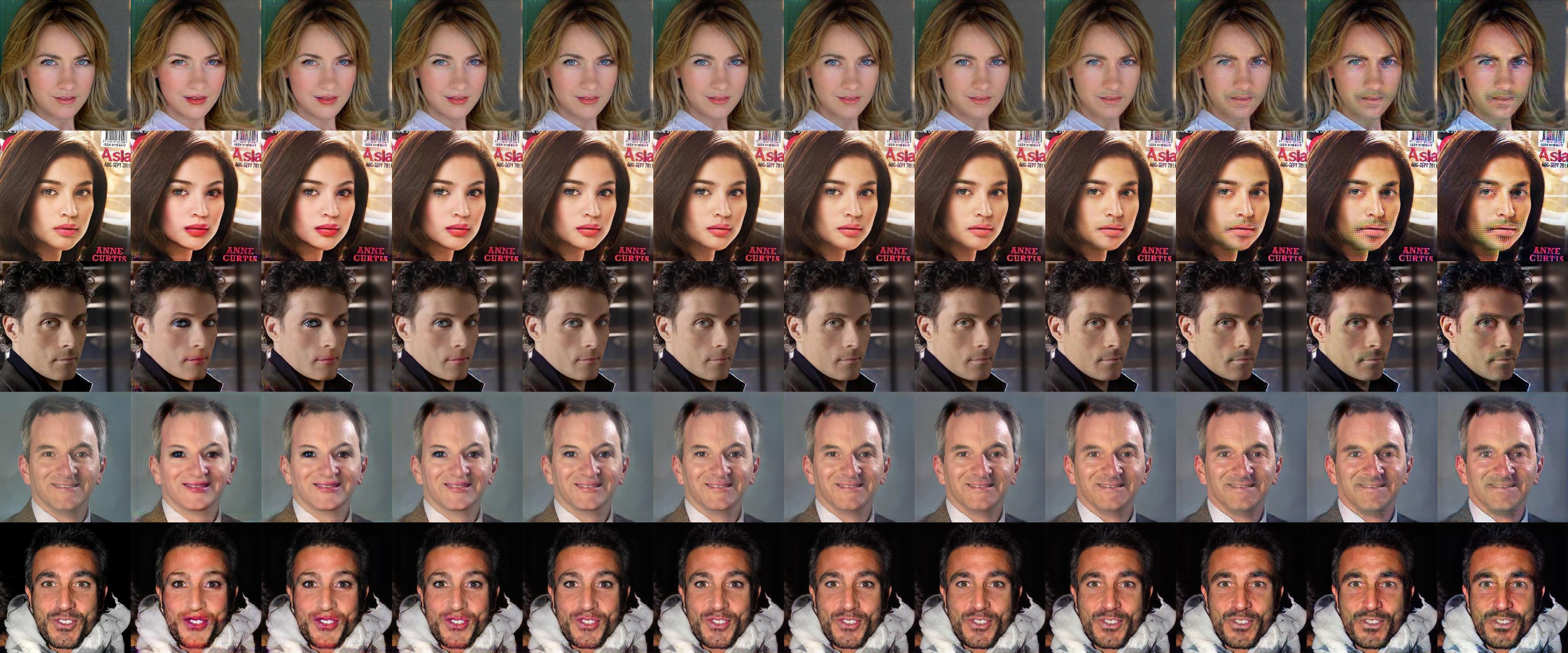}}
 	\centerline{TRIP}
	\end{minipage}	
	\vskip-0.1in
	 \caption{\label{fig:celebahq_male_} The fine-grained facial attribute (``male'') translation on CelebA-HQ dataset by RCGAN and TRIP. The first column is the input image. The other columns are the generated images conditioned on the latent variable from $-1$ to $1$ with step 0.2. The presented results are randomly sampled from the test set.}
\end{figure*}
\end{landscape}

\begin{landscape}
\begin{figure*}[!htb]
\vskip-0.3in
\centering
	\begin{minipage}{0.9\linewidth}
    \centerline{\includegraphics[width=0.98\textwidth]{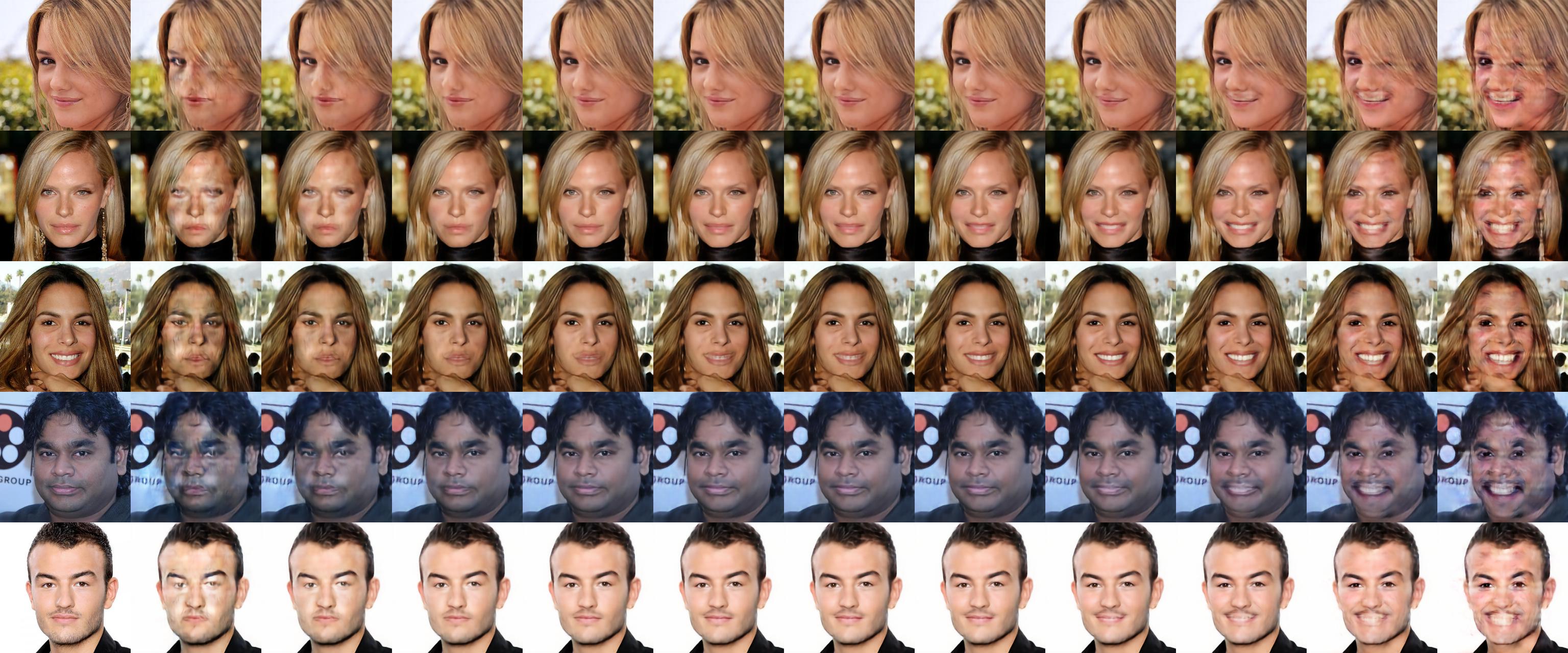}}
	\centerline{FN}
	\end{minipage}
	\\
	\begin{minipage}{0.9\linewidth}
	\centerline{\includegraphics[width=0.98\textwidth]{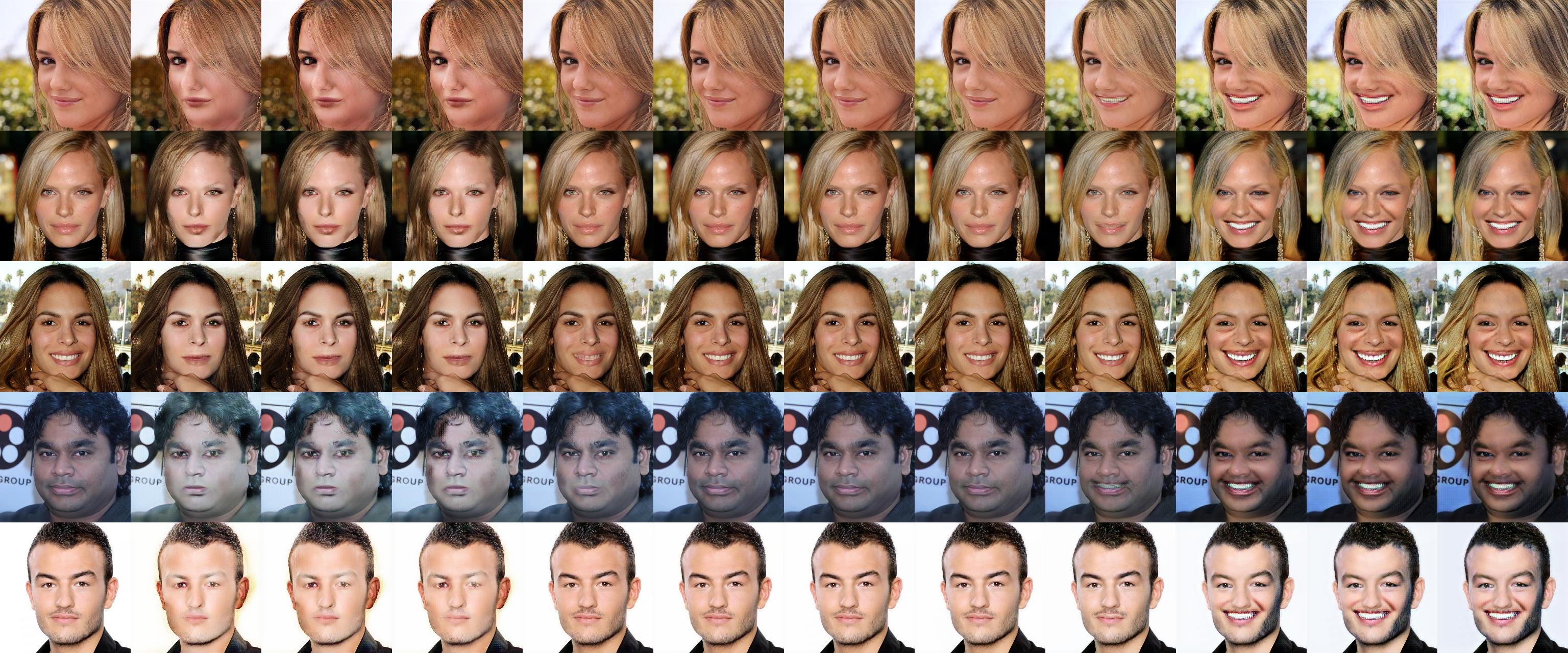}}
	\centerline{RelGAN}
	\end{minipage}
	\vskip-0.1in
	  \caption{ \label{fig:celebahq_smile}The fine-grained facial attribute (``smile'') translation on CelebA-HQ dataset by FN and RelGAN. The first column is the input image. The other columns are the generated images conditioned on the latent variable from $-1$ to $1$ with step 0.2. The presented results are randomly sampled from the test set.}
\end{figure*}
\end{landscape}

\begin{landscape}
\begin{figure*}[!htb]
\vskip-0.3in
\centering
	\begin{minipage}{0.9\linewidth}
 \centerline{\includegraphics[width=0.98\textwidth]{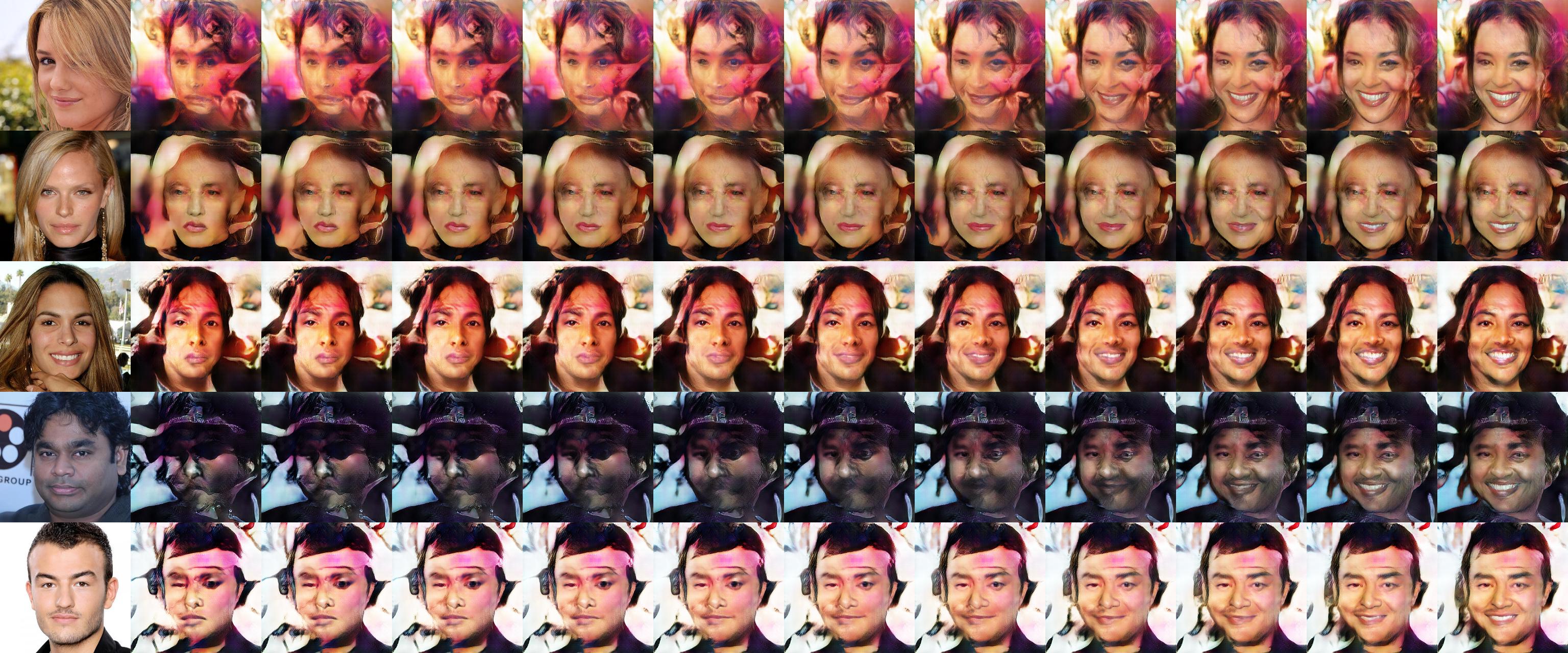}}
 	\centerline{RCGAN}
	\end{minipage}	
	\\
	\begin{minipage}{0.9\linewidth}
 \centerline{\includegraphics[width=0.98\textwidth]{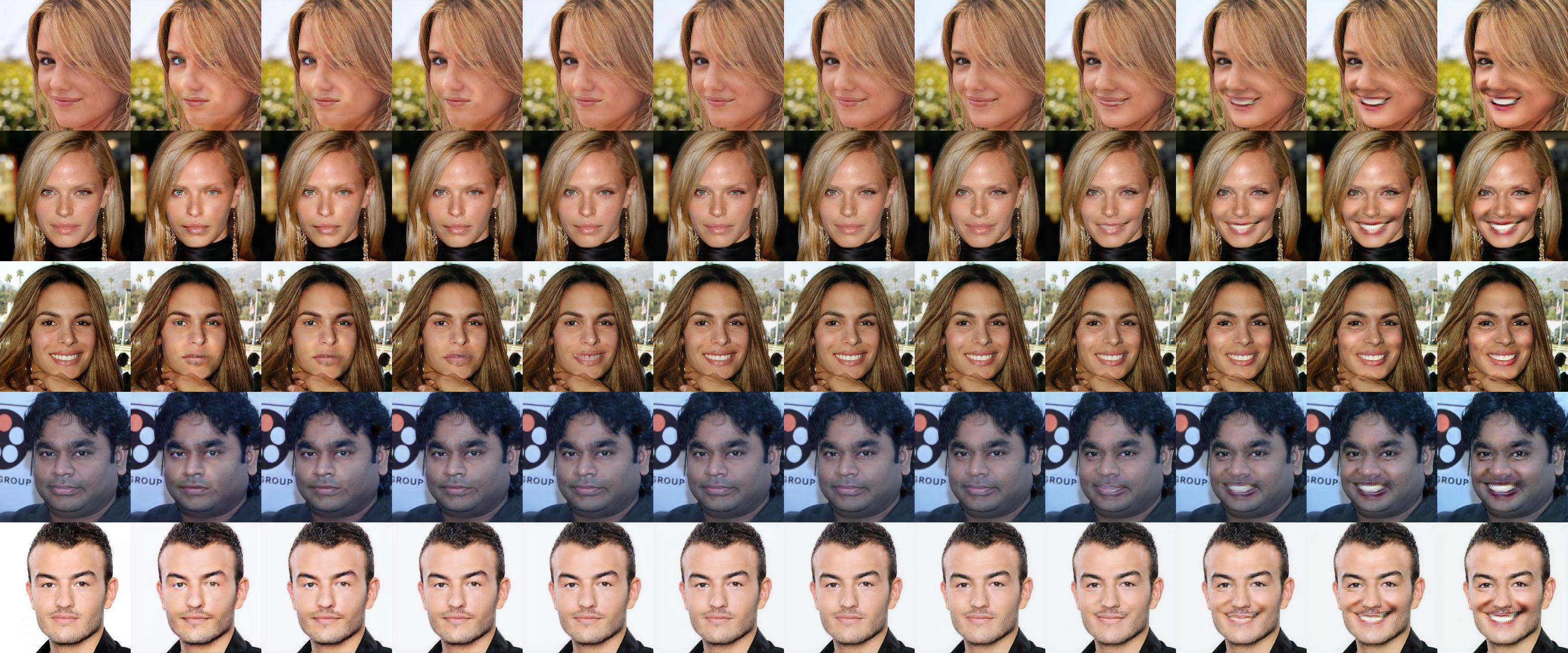}}
 	\centerline{TRIP}
	\end{minipage}	
	\vskip-0.1in
	  \caption{ \label{fig:celebahq_smile_}The fine-grained facial attribute (``smile'') translation on CelebA-HQ dataset by RCGAN and TRIP. The first column is the input image. The other columns are the generated images conditioned on the latent variable from $-1$ to $1$ with step 0.2. The presented results are randomly sampled from the test set.}
\end{figure*}
\end{landscape}

\begin{landscape}
\begin{figure*}[!htb]
\vskip-0.3in
\centering
	\begin{minipage}{0.9\linewidth}
\centerline{\includegraphics[width=0.98\textwidth]{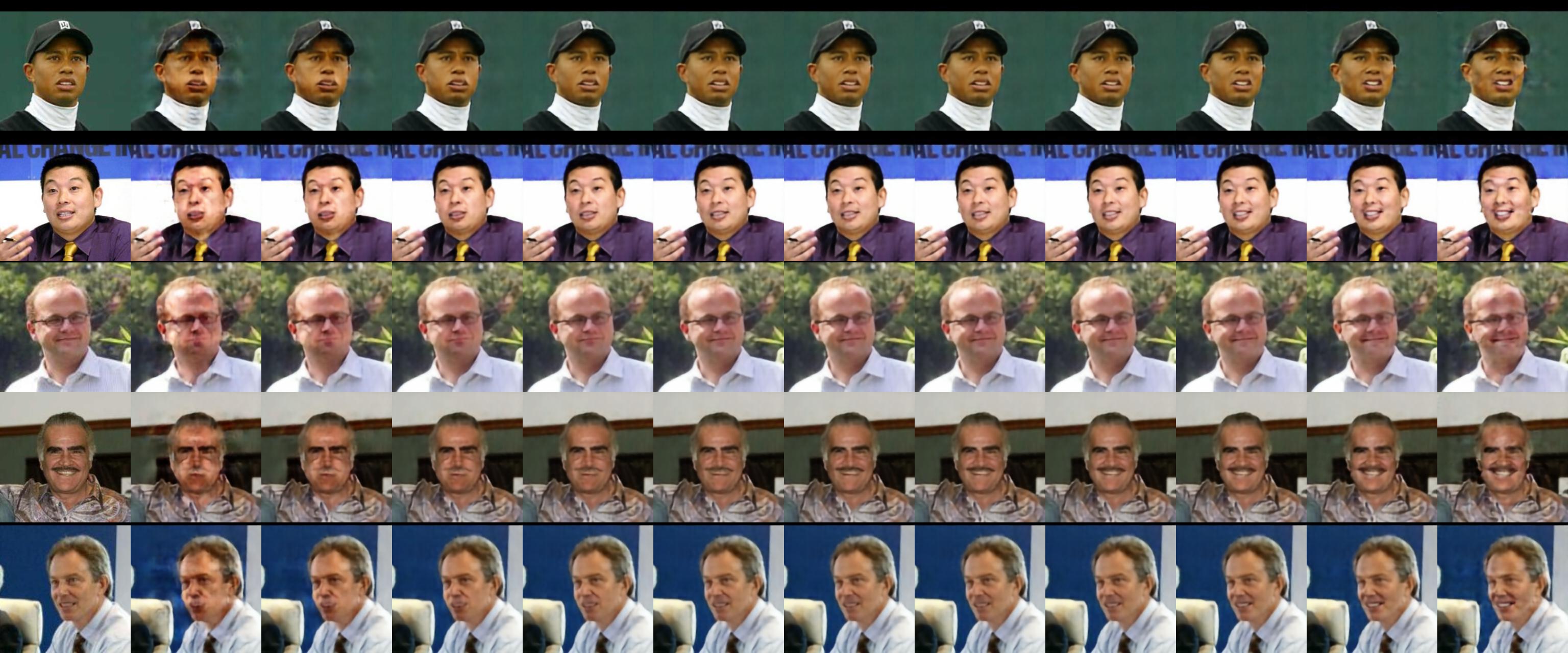}}
	\centerline{FN}
	\end{minipage}
	\\
	\begin{minipage}{0.9\linewidth}
	\centerline{\includegraphics[width=0.98\textwidth]{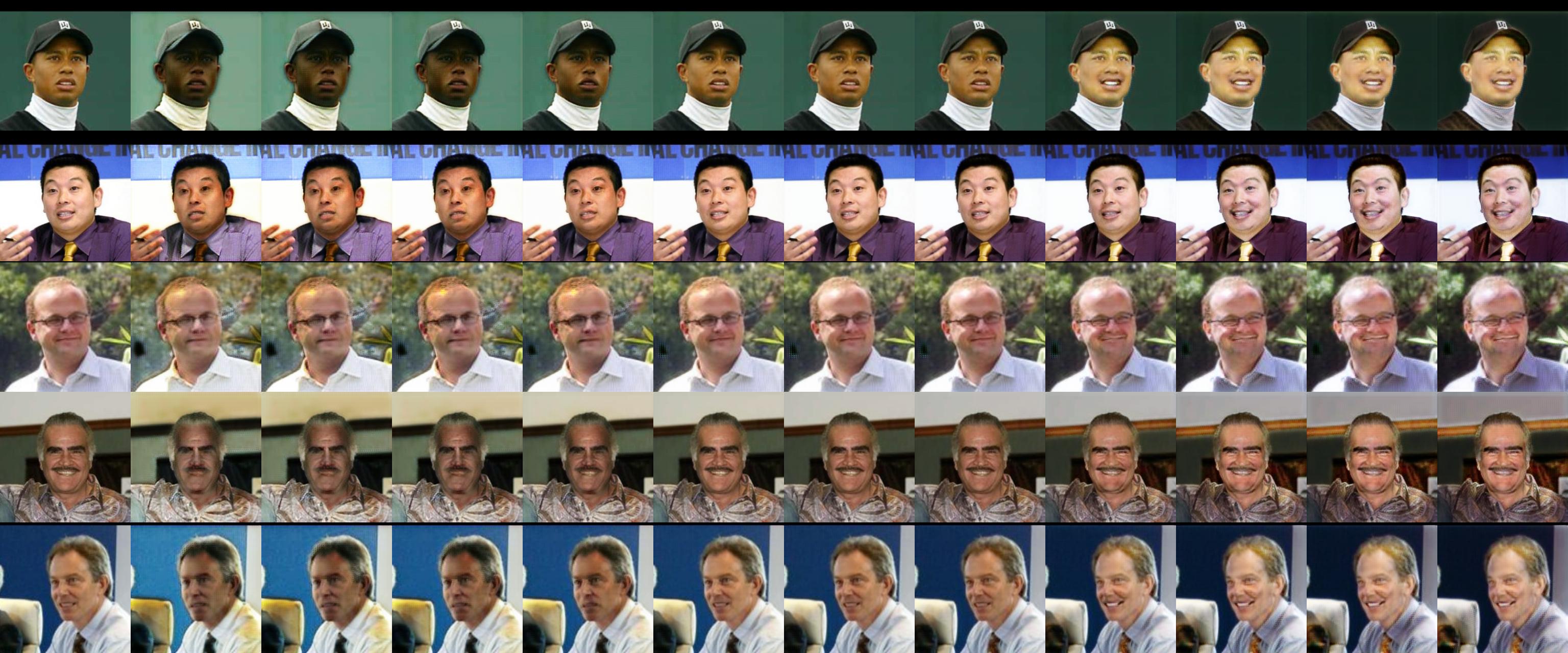}}
	\centerline{RelGAN}
	\end{minipage}
	\vskip-0.1in
	   \caption{\label{fig:lfw_smile} The fine-grained facial attribute (``smile'') translation on LFW dataset by FN and RelGAN. The first column is the input image. The other columns are the generated images conditioned on the latent variable from $-1$ to $1$ with step 0.2. The presented results are randomly sampled from the test set.}
\end{figure*}
\end{landscape}

\begin{landscape}
\begin{figure*}[!htb]
\vskip-0.3in
\centering
	\begin{minipage}{0.9\linewidth}
 \centerline{\includegraphics[width=0.98\textwidth]{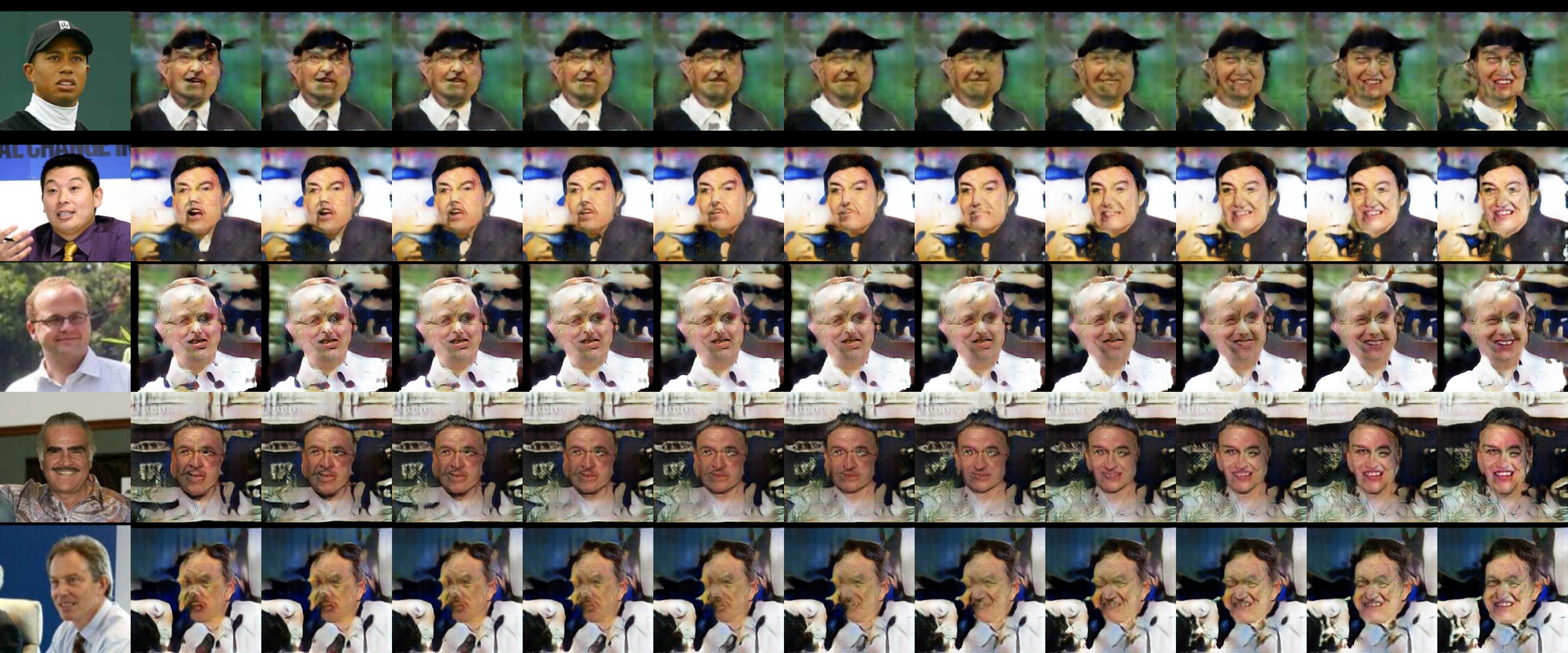}}
 	\centerline{RCGAN}
	\end{minipage}	
	\\
	\begin{minipage}{0.9\linewidth}
 \centerline{\includegraphics[width=0.98\textwidth]{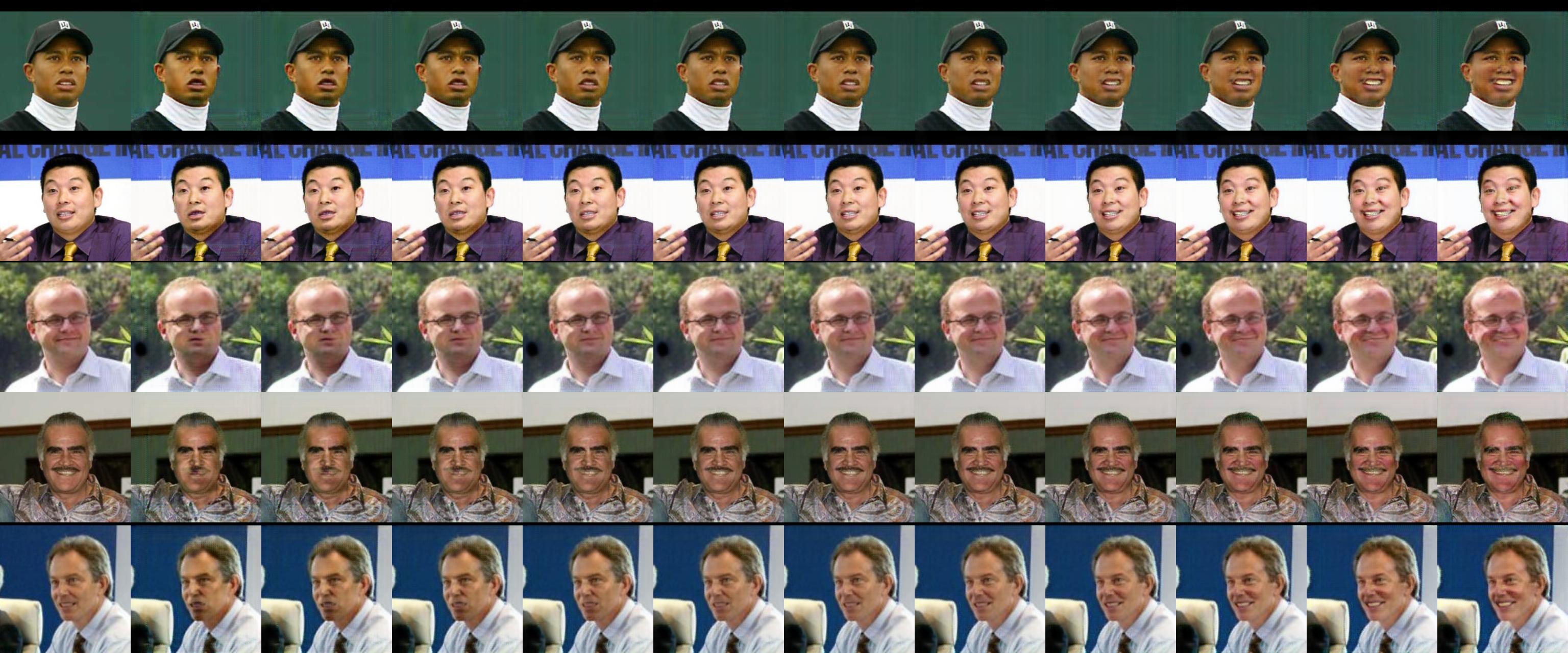}}
 	\centerline{TRIP}
	\end{minipage}	
	\vskip-0.1in
	   \caption{\label{fig:lfw_smile_} The fine-grained facial attribute (``smile'') translation on LFW dataset by RCGAN and TRIP. The first column is the input image. The other columns are the generated images conditioned on the latent variable from $-1$ to $1$ with step 0.2. The presented results are randomly sampled from the test set.}
\end{figure*}
\end{landscape}



\ifCLASSOPTIONcaptionsoff
  \newpage
\fi

\end{document}